\documentclass[table]{bmvc2k}

\usepackage{amsmath}
\usepackage{amssymb}
\usepackage{gensymb}
\usepackage{comment}
\usepackage{booktabs,multirow}
\usepackage{bm}
\usepackage{listings}
\usepackage{float}
\usepackage{array}

\usepackage[subtle]{savetrees}

\usepackage{chngcntr}

\definecolor{codegreen}{rgb}{0,0.6,0}
\definecolor{codegray}{rgb}{0.5,0.5,0.5}
\definecolor{codepurple}{rgb}{0.58,0,0.82}
\definecolor{backcolour}{rgb}{0.95,0.95,0.92}

\lstdefinestyle{mystyle}{
    backgroundcolor=\color{backcolour},   
    commentstyle=\color{codegreen},
    keywordstyle=\color{magenta},
    numberstyle=\tiny\color{codegray},
    stringstyle=\color{codepurple},
    basicstyle=\ttfamily\footnotesize,
    breakatwhitespace=false,         
    breaklines=true,                 
    captionpos=b,                    
    keepspaces=true,                 
    numbers=left,                    
    numbersep=5pt,                  
    showspaces=false,                
    showstringspaces=false,
    showtabs=false,                  
    tabsize=2
}

\newcommand{\keypoint}[1]{\noindent\textbf{#1:}}
\newcommand{\cut}[1]{}

\definecolor{random}{HTML}{ffe0ff}
\definecolor{supervised}{HTML}{ffffe0}
\definecolor{default}{HTML}{e9e9ff}
\definecolor{spatial}{HTML}{e9ffe9}
\definecolor{spatial_dark}{HTML}{c9efc9}
\definecolor{appearance}{HTML}{fff0e3}
\definecolor{appearance_dark}{HTML}{ffe0c3}
\definecolor{grey}{HTML}{dddddd}
\newcommand{\doublecheck}[1]{\textcolor{blue}{#1}}
\newcommand{\todo}[1]{\textcolor{red}{(#1)}}

\usepackage[capitalize]{cleveref}
\crefname{section}{Sec.}{Secs.}
\Crefname{section}{Section}{Sections}
\Crefname{table}{Table}{Tables}
\crefname{table}{Tab.}{Tabs.}

\newcolumntype{g}{>{\columncolor{spatial}}c}
\newcolumntype{o}{>{\columncolor{appearance}}c}

\usepackage{tikz}
\usepackage{comment}
\usepackage{color}

\usepackage[accsupp]{axessibility}  

\title{Why Do Self-Supervised Models Transfer? On \\ the Impact of Invariance on Downstream Tasks}

\addauthor{Linus Ericsson}{linus.ericsson@ed.ac.uk}{1}
\addauthor{Henry Gouk}{henry.gouk@ed.ac.uk}{1}
\addauthor{Timothy M. Hospedales}{t.hospedales@ed.ac.uk}{1, 2}

\addinstitution{
 University of Edinburgh
}
\addinstitution{
 Samsung AI Research, Cambridge
}

\runninghead{Ericsson, Gouk, Hospedales}{Why Do Self-Supervised Models Transfer?}

\begin{document}

\maketitle

\begin{abstract}
Self-supervised learning is a powerful paradigm for representation learning on unlabelled images. A wealth of effective new methods based on instance matching rely on data-augmentation to drive learning, and these have reached a rough agreement on an augmentation scheme that optimises popular recognition benchmarks. However, there is strong reason to suspect that different tasks in computer vision require features to encode different (in)variances, and therefore likely require different augmentation strategies. In this paper, we measure the invariances learned by contrastive methods and confirm that they do learn invariance to the augmentations used and further show that this invariance largely transfers to related real-world changes in pose and lighting. We show that learned invariances strongly affect downstream task performance and confirm that different downstream tasks benefit from polar opposite (in)variances, leading to performance loss when the standard augmentation strategy is used. Finally, we demonstrate that a simple fusion of representations with complementary invariances ensures wide transferability to all the diverse downstream tasks considered.
\end{abstract}

\section{Introduction}
Self-supervised learning has made rapid progress in representation learning, with performance approaching and sometimes surpassing that of supervised pre-training. In computer vision contrastive self-supervised methods driven by data augmentation have been particularly effective \cite{He2019MomentumLearning,Chen2020BigLearners}. Data augmentation applies synthetic semantics-preserving transformations to images during learning, to increase effective data volume and promote invariance to the augmentation distribution used \cite{Shorten2019ALearning,Cubuk2020RandAugment:Space}. By optimising representations so that individual images are similar to their augmented counterparts \cite{Grill2020BootstrapLearning,Chen2021ExploringLearning}, and possibly also different to alternative distractor images \cite{He2019MomentumLearning,Chen2020BigLearners,Wang2020UnderstandingHypersphere}, self-supervised algorithms have achieved wide success \cite{Ericsson2021HowTransfer}.

\begin{figure}[t]
    \centering
    \includegraphics[width=0.72\linewidth]{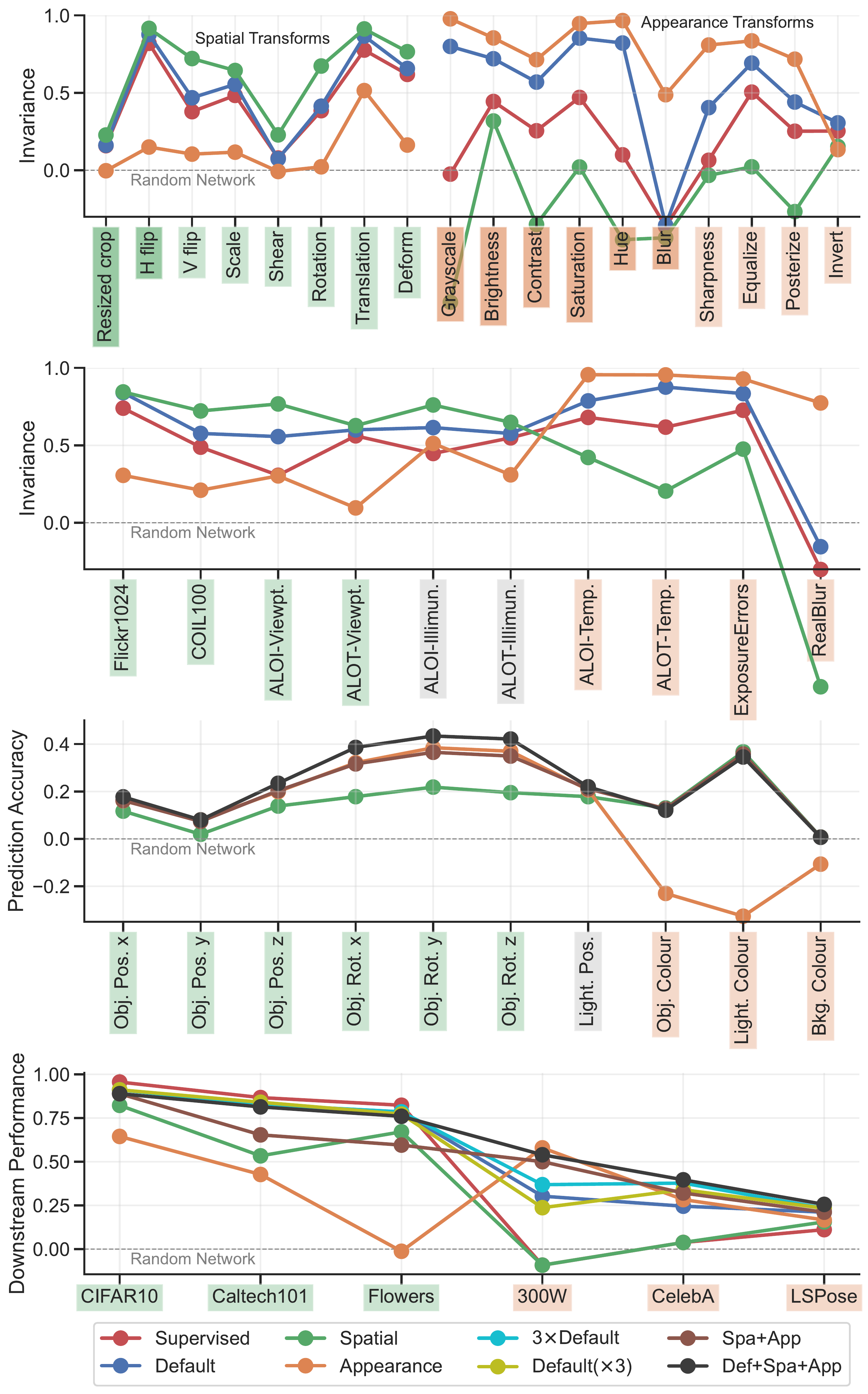}
    \caption{Our Spatial and Appearance models lead to strong spatial and colour/texture invariance respectively, as measured by both synthetic (first row) and real-world (second row) transforms. Simple feature fusion (black) dominates individual pathways, as well as state of the art `default' augmentation, providing more consistent performance across all downstream tasks (third and fourth row).}
    \label{fig:teaser}
\end{figure}

In this paradigm the properties and efficacy of the learned representation are largely determined by the augmentation distribution used during self-supervision. To this end a rough consensus has emerged among many state of the art methods as to a good default distribution that leads to strong performance on the downstream benchmarks, especially on the ubiquitous ImageNet object recognition benchmark \cite{Deng2009ImageNet:Database}. For example, image cropping, flipping, colour perturbation and blurring, are widely applied \cite{He2019MomentumLearning,Chen2020ImprovedLearning,Ericsson2021HowTransfer}. However, if augmentation leads to invariance to the corresponding transformation, then we should ask: do our self-supervised algorithms provide the right invariances for diverse downstream tasks of interest? For example, while an object categorisation task might benefit from pose invariance, 
other tasks such as pose estimation may require strong spatial sensitivity. If different tasks require contradictory (in)variances, using a single default data augmentation scheme for all may provide sub-optimal performance for some tasks. 

To investigate this issue, we group augmentations into two categories, 
\textit{spatial} 
and 
\textit{appearance}
. Using a representative state of the art contrastive learner MoCo-v2+ResNet50 \cite{Chen2020ImprovedLearning}, we train models exclusively with spatial-style and appearance-style augmentations and compare them to the model produced by the default augmentation scheme. In particular, we evaluate their resulting invariances to synthetic and real-world transforms, as well as their performance on a suite of diverse real-world downstream tasks.

Based on the experimental design outlined above, we attempt to better understand \emph{why contrastive self-supervised learning works} by answering the following specific questions, among others, with associated results summarised in Figure~\ref{fig:teaser}.

\textbf{Q1} \emph{An increasing amount of work has shown that invariances can be learned by learning augmentations. Do these learned synthetic invariances generalise to real-world invariances?} A1: To some degree, yes. For example, spatial-style augmentations lead to increased invariance to real-world transforms such as viewpoint, while appearance-style augmentations lead to increased invariance to transformations such as lighting colour, exposure and blur. Correspondingly, spatial-style augmentations lead to higher accuracy in estimating object colour, while appearance-style augmentations lead to higher accuracy in estimating object pose. (Fig.~\ref{fig:teaser} second and third row). This has not been measured before.

\textbf{Q2} \emph{Given that there are multiple types invariances of potential interest to learn. Is there a trade-off between learning different types of invariances?} A2: Yes. Promoting appearance-style invariances decreases spatial-style ones and vice-versa. We also show that all existing state-of-the-art learners suffer from this trade-off.

\textbf{Q3} \emph{Do different downstream tasks of interest benefit from different invariances?} A3: Yes. Across a suite of downstream tasks, we see that recognition-style tasks prefer a representation trained on default or spatial-style augmentations, while pose-related tasks benefit from appearance-style augmentations. In particular, default augmentations \cite{Chen2020ImprovedLearning} under-perform in pose-related tasks (Fig.~\ref{fig:teaser} fourth row). 

\textbf{Q4} \emph{Given that different tasks prefer polar-opposite augmentations, is there a simple way to achieve high performance across all tasks?} A4: Yes. Simple fusion of multiple representations tuned for different (in)variances leads to consistent strong performance across all tasks considered (Fig.~\ref{fig:teaser} third and third fourth row, black line).

\section{Related Work}
\keypoint{Self-supervision} in computer vision is now too a large topic to review here. Please see \cite{Jing2020Self-supervisedSurvey,ericsson2021self} for excellent surveys. A key trend is that many highly successful methods rely on matching individual images with augmented versions of themselves, possibly against a background of distractor images. This includes most contrastive methods \cite{Chen2020ARepresentations,He2019MomentumLearning}, and some that are not typically considered contrastive \cite{Zbontar2021BarlowReduction,Grill2020BootstrapLearning}. These have been understood \cite{Wang2020UnderstandingHypersphere} as making image features invariant to transformations used for training, while otherwise separating individual images. 
A key ambition of self-supervision research is for a single pre-trained feature to support diverse downstream tasks, and a common suite of augmentations has emerged to support this \cite{Chen2020ImprovedLearning,Ericsson2021HowTransfer}. However if data augmentation determines (in)variances, and different tasks require different (in)variances, then a single augmentation distribution may not perform well on all tasks.

\cut{\keypoint{Building in Invariances}
Early work in feature engineering produced SIFT and SURF features that were designed to be invariant to various transformations \cite{YounusKhan2011SIFTDataset}. With the advent of deep learning, the focus shifted to enforcing feature properties via neural network architectures. For images, convolutional neural networks \cite{Fukushima1980Neocognitron:Position,Krizhevsky2012ImageNetNetworks,He2016DeepRecognition} have long been the go-to architecture because of their weight-sharing and spatial equivariance properties. Many other architectures have been developed that capture different or more in-/equi-variances \cite{Cohen2016GroupNetworks,Cohen2019GaugeCNN,Sabour2017DynamicCapsules}. We focus on understanding invariances due to learning under data augmentation instead.}

\keypoint{Data Augmentation}
Data augmentation is the process of transforming input data to increase the diversity of the training set. It has become key to achieving state-of-the-art performance for supervised learning of CNNs in  vision  \cite{Cubuk2019AutoAugment:Data}. 
Despite its ubiquity in practice, theoretical understanding of data augmentation is weak.
There is some evidence that CNNs can generalise learned translation invariance \cite{Biscione2020LearningCNNs} to unseen data, but also that they retain information about absolute spatial locations of objects via boundary effects \cite{Kayhan2020OnLocation}.
\cut{and feature invariances have been studied theoretically by comparing data augmentation or feature averaging of augmented inputs \cite{Lyle2020OnNetworks}.}

Data augmentation has become even more vital in practical self-supervision as outlined earlier. However, understanding the role of data augmentation in self-supervision has lagged behind practical engineering lore. Self-supervised contrastive learners with strong augmentation have been shown to learn occlusion-invariant representations, but not to capture viewpoint and category instance invariance~\cite{Purushwalkam2020DemystifyingBiases}. \cite{Wen2021TowardLearning} study the theoretical effects of data augmentation on self-supervised contrastive learning. They argue that data augmentation decouples sparse semantic information in the input from dense noisy information and that only the sparse semantic information is relevant to solving the downstream target task. 
\cite{vonKugelgen2021Self-SupervisedStyle} study the effects of data augmentation on invariances and downstream performances using a synthetic task. They show that it is hard to define augmentations to enforce a specific invariance, that augmentations generally have wider invariance effects on groups of factors and that using multiple augmentations in conjunction reliably improves recognition performance. However they focus on object recognition in a synthetic dataset. We take a wider perspective and look at how augmentation impacts a wide variety of real-world transformations, and real-world downstream tasks.  
A related study to ours is \cite{Xiao2021WhatLearning}, which proposes LooC as a self-supervised method that separates different information into different features, i.e. colour, orientation etc. However, they only evaluate the impact on recognition tasks. A major contribution of ours is to demonstrate how diverse downstream tasks benefit from different learned invariances.

\keypoint{Ventral-Dorsal Visual System} We also briefly highlight an interesting connection between our spatial vs.~appearance split and neuroscience. A well established theory about mammalian vision holds that the visual cortex is composed of two functional pathways \cite{Goodale1992SeparateAction,kruger2013deepHierarchies}. The \textit{ventral} stream deals with the ``what'' of object recognition; and the \textit{dorsal} stream deals with the ``where'' of spatial and motion information. This decomposition into specialised models has been exploited in applications such as object detection \cite{Ebrahimpour2019Ventral-DorsalAttention} and semantic grasping \cite{Jang2017End-to-EndGrasping} in robotics. At the intersection of neuroscience and self-supervised learning, \cite{Bakhtiari2021TheLearning} showed that a two branch neural network trained with the CPC \cite{Oord2017NeuralLearning} loss on videos leads to dorsal and ventral-like pathways emerging. Moreover, models of the dorsal stream based entirely on findings from neuroscience and phychophysics (i.e., without use of machine learning) have been shown to accurately estimate motion and depth from videos \cite{cree2016estimating,perrone2019using}.

We explore self-supervised learning with different data augmentations as a way of achieving similar multi-stream pathways in CNNs for vision. Current methods \cite{Chen2020ARepresentations,He2019MomentumLearning} train representations for invariance to a single set of augmentations that aim to suffice for all tasks. But we show that this current practice is better optimised for the most popular downstream benchmark of object recognition, and poor for pose-related tasks. We will investigate a multi-stream architecture combining representations trained for different invariances, and show it provides more general purpose high performance for diverse downstream tasks.


\section{Methods}
Our main focus is on analysing the properties of self-supervised models pre-trained with different augmentation strategies. 
In particular, we choose MoCo-v2 \cite{Chen2020ImprovedLearning} as a representative self-supervised learner that is widely used and near state-of-the-art. MoCo-v2 matches images with their augmented counterparts, while using negative pairs in a contrastive loss to encourage feature dissimilarity between semantic objects, and to avoid features all collapsing to the same vector. 
\cut{We pre-train models using one of these regularisation methods in addition to the contrastive loss of MoCo-v2. Additionally it would be interesting to compare a model pre-trained with a supervised objective. However, at this stage we are only able to present results for MoCo-v2 pre-trained models.}
We pre-train three models using MoCo-v2 \cite{Chen2020ImprovedLearning} with ResNet50 architectures \cite{He2016DeepRecognition} on ImageNet \cite{Deng2009ImageNet:Database} for 200 epochs.

\begin{itemize}
    \item \keypoint{Default} The default \cite{Chen2020ImprovedLearning,Chen2020ARepresentations,Chen2020BigLearners,Li2021PrototypicalRepresentations,Grill2020BootstrapLearning} model uses the standard array of data augmentations, which includes crops, horizontal flips, color jitter, grayscale and blur.
    \item \keypoint{Spatial} The Spatial model uses only the spatial subset of default augmentations, including crops and horizontal flips. By learning invariance to these spatial transforms, the model has to put larger focus on colour and texture.
    \item \keypoint{Appearance} The Appearance model uses only the appearance-based augmentations of color jitter, grayscale and blur and will thus have to put larger focus on spatial information.
\end{itemize}

\begin{table}[t]
    \centering
    \caption{Augmentations used during pre-training of our Spatial and Appearance models, along with the standard default augmentations \cite{Chen2020ImprovedLearning}. The color jitter augmentation is a combination of individual jitter in brightness, contrast, saturation and hue.}
    \label{tab:pretrain_model_augmentations}
    \resizebox{0.6\linewidth}{!}{%
    \begin{tabular}{r|ccccc}
        \toprule
                   & \cellcolor{spatial_dark}Resized crop       & \cellcolor{spatial_dark}Horizontal flip     & \cellcolor{appearance_dark}Color jitter     & \cellcolor{appearance_dark}Grayscale     & \cellcolor{appearance_dark}Blur \\
        \midrule
        Default    & \checkmark & \checkmark & \checkmark & \checkmark    & \checkmark \\
        \cellcolor{spatial_dark}Spatial    & \checkmark & \checkmark &            &               &            \\
        \cellcolor{appearance_dark}Appearance     &            &            & \checkmark & \checkmark    & \checkmark \\
        \bottomrule
    \end{tabular}
    }
\end{table}

Table \ref{tab:pretrain_model_augmentations} summarises the augmentations used by each model. Apart from these differences, the pre-training setup is identical for our models. As baselines, we also compare a CNN with \textbf{Random} weights, and one pre-trained by \textbf{Supervised} learning on ImageNet.

\cut{
\subsection{Feature Averaging} The models above aim to learn invariances through contrastive learning with data augmentation. An alternative strategy \cite{ashukha2021testAugment} to realising invariance with attractive theoretical properties \cite{Lyle2020OnNetworks} is test-time feature averaging (FA). Given any feature extractor $f$ and an image $x$, we can define the mean feature $\bar{z}$ under a transform function $t$ as
\begin{equation}
    \label{eq:feature_averaging_expectation}
    \bar{z} =  \mathbb{E}_{x' \sim t(x)} \Big[ f(x') \Big].
\end{equation}
It is easy to see that for some cases such as a uniform distribution over rotations, this embedding is completely rotation invariant, independent of the underlying feature extractor $f(\cdot)$ properties. Our FA models estimate this vector by sampling $k = 32$ augmented images, passing them through the backbone before the resulting features are averaged into a single feature vector to represent the input. The FA strategy can be applied upon any base model, and we choose to apply it to a pre-trained supervised model. Our feature-averaging Spatial model (FA-Spatial) uses the same augmentations of resized crop and horizontal flip as the Spatial model uses during pre-training. Similarly, our FA-Appearance model uses feature averaging with color jitter, grayscale and blurring.}


\section{Do Contrastive Methods Learn Invariance to Real-World Transforms?}
\label{sec:synthetic-invariance}

While several preliminary studies suggest that contrastive methods can learn invariance to synthetic transformations \cite{Wang2020UnderstandingHypersphere}, an important question that has not been studied in the literature is whether these learned invariances lead to invariance under real-world transforms, like viewpoint or illumination changes. Does the use of colour augmentations during pre-training lead to features that are invariant to day/night in real images? Does the use of crop/flip augmentation in training lead to pose invariance in real images? While these kinds of questions were intensively studied for classic hand-crafted features \cite{mikolajczyk2005descriptor_eval}, they have not been studied for invariances learned by self-supervision. In this section, we investigate whether contrastive methods learn invariance to real-world transforms. We address this question from two perspectives: intrinsically, by measuring the invariance of different representations with respect to different real-world transformations (Section~\ref{sub:intrinsic}); and extrinsically, by quantifying how well features trained for different synthetic invariances can be used to predict known real-world transformations (Section~\ref{sub:extrinsic}). But first, we provide a more thorough confirmation of the claim that contrastive methods do learn invariances to synthetic transformations.

\keypoint{Measuring invariances}
We use two measures of invariance in our experiments, Mahalanobis distance and cosine similarity (full details in Sec~\ref{sec:measure} of supplement). We compute these values between augmented and unaugmented images, averaged over all images considered. A further set of measures are reported in Sec~\ref{sec:alt_measure} of the supplement with results supporting those in the main paper.

\keypoint{Hypothesis testing}
In the following sections we will test several hypotheses based on the cosine similarity invariance measurements. We make use of Hoeffding's inequality, which for a sum of random variables, $S_n = \frac{1}{n}\sum_{i=1}^n X_i$, where each $X_i$ is in the range $[0, 1]$ with probability one, tells us that
\begin{equation}
    P(S_n - \mathbb{E}[S_n] \geq t) \leq e^{-2nt^2}.
\end{equation}
Setting the left-hand side equal to $\delta$ and rearranging for $t$ yields
\begin{equation}
    t \leq \sqrt{\frac{\text{ln}(1/\delta)}{2n}}.
    \label{eq:threshold}
\end{equation}
This fact can be used to test the null hypothesis that the expected value of $S_n$ is zero: set $\delta$ to the threshold that will be applied to a p-value, and check whether $S_n$ is greater than the right-hand side of Eq. \ref{eq:threshold}. If it is greater, then one can reject the null hypothesis. By setting $S_n$ equal to the mean difference in representation similarity for two different methods, we can test whether one method is statistically significantly more invariant than the other. Bonferroni correction is applied when we carry out multiple hypothesis tests to perform a three-way comparison.

\subsection{Invariance to Synthetic Transforms}\label{sec:synthetic}
\keypoint{Setup} We focus on task-agnostic metrics of invariances. Other extrinsic measures of invariance like identifiability/classification performance under different transformations are inherently biased towards that task. We therefore use invariance metrics that apply to feature vectors directly. We evaluate our Default, Spatial and Appearance methods on 1,000 images from the ImageNet (ILSVRC12) validation set  \cite{Deng2009ImageNet:Database} against a wider array of synthetic augmentation transformations than used for training (Tab~\ref{tab:pretrain_model_augmentations}), but still group these into appearance and spatial-style transforms.

\begin{table*}[t]
    \centering
    \caption{ImageNet pre-trained ResNet50 with MoCo-v2 (200 epochs) evaluated on invariances to transforms on 1000 ImageNet validation images. Top group: Mahalanobis distance where a low value means strong invariance. Bottom group: cosine similarity in a normalised feature space where a value close to 1 means strong invariance. Column colours indicate the type of invariance evaluated and row colours indicate the augmentation expected to lead to high-performing specialised models. The broad agreement between the most invariant features (bold) and expectation (row colours) indicates that training with augmentations does tend to learn the corresponding invariances. Similarity results within \{Default, Spatial, Appearance\} that are statistically significantly the best are annotated with a $\bullet$.}
    \label{tab:synthetic_invariances}
    \resizebox{1.\textwidth}{!}{%
    \begin{tabular}{lllcccccccccccccccccc}
    \toprule
    & & {} &  \cellcolor{spatial_dark}Resized crop &  \cellcolor{spatial_dark}H flip &  \cellcolor{spatial}V flip &  \cellcolor{spatial}Scale &  \cellcolor{spatial}Shear &  \cellcolor{spatial}Rotation &  \cellcolor{spatial}Translation &  \cellcolor{spatial}Deform &  \cellcolor{appearance_dark}Grayscale &  \cellcolor{appearance_dark}Brightness &  \cellcolor{appearance_dark}Contrast &  \cellcolor{appearance_dark}Saturation &  \cellcolor{appearance_dark}Hue &  \cellcolor{appearance_dark}Blur &  \cellcolor{appearance}Sharpness &  \cellcolor{appearance}Equalize &  \cellcolor{appearance}Posterize &  \cellcolor{appearance}Invert \\
    \midrule
    \multirow{5}{*}{\rotatebox{90}{Distance}} & \multirow{5}{*}{$\downarrow$} & Random     &         69.40 &   34.14 &   35.35 &  67.03 &  69.57 &     71.65 &        56.33 &   65.28 &      22.81 &       73.25 &     59.03 &       46.63 & 42.39 & \underline{49.17} &      52.59 &     27.46 &      27.46 &   32.88 \\
    & & Supervised &         \textbf{57.44} &   12.33 &   24.07 &  40.37 &  63.93 &     47.67 &        22.51 &   34.25 &      19.87 &       40.43 &     37.05 &       26.61 & 35.95 & 54.92 &      42.15 &     17.44 &      22.74 &   \underline{27.32} \\
    & & Default    &         \underline{58.72} &    \underline{9.92} &   \underline{21.30} &  \underline{35.75} &  \underline{56.58} &     \underline{43.49} &        \underline{16.45} &   \underline{30.16} &       \underline{7.78} &       \underline{26.07} &     \underline{25.66} &       \underline{12.17} & \underline{13.72} & 65.84 &      \underline{31.07} &     \underline{12.81} &      \underline{17.95} &   \textbf{24.71} \\
    & & \cellcolor{spatial_dark}Spatial    &         \cellcolor{spatial}59.43 &    \cellcolor{spatial}\textbf{8.17} &   \cellcolor{spatial}\textbf{15.57} &  \cellcolor{spatial}\textbf{32.05} &  \cellcolor{spatial}\textbf{56.50} &     \cellcolor{spatial}\textbf{32.93} &        \cellcolor{spatial}\textbf{13.85} &   \cellcolor{spatial}\textbf{26.08} &      26.58 &       46.25 &     61.67 &       39.34 & 47.33 & 63.37 &      48.83 &     25.41 &      38.46 &   28.95 \\
    & & \cellcolor{appearance_dark}Appearance     &         64.35 &   27.52 &   29.05 &  56.33 &  71.81 &     62.49 &        33.14 &   52.20 &       \cellcolor{appearance}\textbf{2.57} &       \cellcolor{appearance}\textbf{19.71} &     \cellcolor{appearance}\textbf{22.84} &        \cellcolor{appearance}\textbf{6.98} &  \cellcolor{appearance}\textbf{5.77} & \cellcolor{appearance}\textbf{30.38} &      \cellcolor{appearance}\textbf{16.86} &      \cellcolor{appearance}\textbf{9.55} &      \cellcolor{appearance}\textbf{12.18} &   \cellcolor{appearance}29.24 \\
    \midrule
    \multirow{5}{*}{\rotatebox{90}{Similarity}} & \multirow{5}{*}{$\uparrow$}  & Random     &          0.03 &    0.56 &    0.54 &   0.16 &   0.04 &      0.07 &         0.40 &    0.20 &       0.81 &        0.17 &      0.52 &        0.59 & 0.60 &  \underline{0.48} &       0.51 &      0.68 &       0.70 &    0.52 \\
    & & Supervised &          0.18 &    0.92 &    0.71 &   0.57 &   \underline{0.11} &      0.43 &         0.87 &    0.69 &       0.81 &        0.54 &      0.64 &        0.79 & 0.64 &  0.29 &       0.55 &      0.84 &       0.77 &    \underline{0.64} \\
    & & Default    &          \underline{0.19} &    \underline{0.95} &    \underline{0.75} &   \underline{0.63} &   \underline{0.11} &      \underline{0.45} &         \underline{0.92} &    \underline{0.72} &       \underline{0.96} &        \underline{0.77} &      \underline{0.79} &        \underline{0.94} & \underline{0.93} &  0.29 &       \underline{0.71} &      \underline{0.90} &       \underline{0.83} &    \textbf{0.67} \\
    & & \cellcolor{spatial_dark}Spatial    &          \cellcolor{spatial}\textbf{0.25}$\bullet$ &    \cellcolor{spatial}\textbf{0.96} &    \cellcolor{spatial}\textbf{0.87}$\bullet$ &   \cellcolor{spatial}\textbf{0.70}$\bullet$ &   \cellcolor{spatial}\textbf{0.26}$\bullet$ &      \cellcolor{spatial}\textbf{0.70}$\bullet$ &         \cellcolor{spatial}\textbf{0.95}$\bullet$ &    \cellcolor{spatial}\textbf{0.81}$\bullet$ &       0.65 &        0.43 &      0.35 &        0.60 & 0.42 &  0.25 &       0.50 &      0.69 &       0.62 &    0.59 \\
    & & \cellcolor{appearance_dark}Appearance     &          0.03 &    0.63 &    0.59 &   0.26 &   0.03 &      0.09 &         0.71 &    0.33 &       \cellcolor{appearance}\textbf{1.00} &        \cellcolor{appearance}\textbf{0.88}$\bullet$ &      \cellcolor{appearance}\textbf{0.86}$\bullet$ &        \cellcolor{appearance}\textbf{0.98} & \cellcolor{appearance}\textbf{0.99}$\bullet$ &  \cellcolor{appearance}\textbf{0.73}$\bullet$ &       \cellcolor{appearance}\textbf{0.91}$\bullet$ &      \cellcolor{appearance}\textbf{0.95}$\bullet$ &       \cellcolor{appearance}\textbf{0.91}$\bullet$ &    \cellcolor{appearance}0.58 \\
    \bottomrule
    \end{tabular}
    }
\end{table*}

\keypoint{Results}
The results in \cref{tab:synthetic_invariances} evaluate the invariance of different transformations at test-time (columns) for the different pre-trained models (rows). Using the method described above, we carry out statistical hypothesis tests to determine which of the Default, Appearance, and Spatial models reliably exhibit the most invariance, as measured by the similarity metric. Statistically significant results (at the 95\% confidence level) are marked with a $\bullet$. We  make the following observations. For spatial transformations like rotation and translation, the Spatial model is the most invariant, due to its use of such augmentations during pre-training. Likewise, the Appearance model has the strongest invariance to transformations in colour and texture, except for the invert transform. The Default model tends to fall in between the two specialised models suggesting strong invariance to any one transformation is traded off for a reasonable variance across the board. The Random model tends to have the highest variance.

While the spatially-augmented model has very low variance to spatial transforms, it has a high variance to colour and texture. Its sensitivity to these transforms is available for solving tasks that depend on colour or texture. Likewise, the appearance-augmented model is sensitive to spatial information which it could use to solve spatially sensitive tasks. In fact, since the Appearance model is more spatially sensitive than the Default model, it might achieve better performance on such tasks. We investigate this in \cref{sec:downstream_tasks}.
Overall the results confirm that invariances are indeed learned by contrastive learning with corresponding augmentations. Furthermore, augmentations do tend to increase invariance to other transforms in the corresponding appearance/spatial family, rather than only the specific subset used for training.

\keypoint{Discussion}
We have shown how the use of certain augmentations lead to features that are substantially invariant to those augmentations, as well as others in the same appearance/spatial family. We next address the main question that has not been studied in the literature of whether these learned invariances lead to invariances under real-world transforms, like viewpoint or illumination changes. 
While these kinds of questions were intensively studied for classic hand-crafted features \cite{mikolajczyk2005descriptor_eval}, they have not been studied for invariances learned by self-supervision.

\subsection{Real-World Intrinsic Invariance Measurements}\label{sub:intrinsic}
\keypoint{Experimental Details} We use the same metrics as in Section \ref{sec:synthetic}, but instead of using synthetic transformations typically used in data augmentation schemes, we collect a suite of datasets that exhibit known real-world transformations, such as pose changes (Flickr1024, COIL-100. ALOI, ALOT) and colour/appearance changes (ALOI, ALOT, ExposureErrors, RealBlur). Further dataset details can be found in the Sec.~\ref{sec:real_world_datasets} of the supplementary materials. In contrast to the experimental setup in Section~\ref{sec:synthetic-invariance}, we do not have an untransformed reference image. Instead, we consider all pairs of images for a given object/scene/texture within each dataset (or subset of the dataset for ALOI/ALOT), and average our metrics across pairs. For example, in the case of viewing angle variation in COIL-100, a fully pose invariant feature would exhibit full similarity/zero distance across all corresponding image pairs.

\begin{table*}[t]
    \centering
    \caption{Comparing models in terms of their invariances to real-world transformations. Similarity results within \{Default, Spatial, Appearance\} that are statistically significantly the best are annotated with a $\bullet$.}
    \label{tab:realworld_invariances}
    \resizebox{1.0\linewidth}{!}{%
    \begin{tabular}{lllccccccccccc}
    \toprule
    {} & {} & {} &  \cellcolor{spatial}Flickr1024 &  \cellcolor{spatial}COIL100 &  \cellcolor{spatial}ALOI &  \cellcolor{spatial}ALOT &  \cellcolor{grey}ALOI &  \cellcolor{grey}ALOT &  \cellcolor{appearance}ALOI &  \cellcolor{appearance}ALOT &  \cellcolor{appearance}ExposureErrors &  \cellcolor{appearance}RealBlur \\
    \midrule
    {} & {} & {} &  \cellcolor{spatial}Stereo &  \cellcolor{spatial}Pose/Scale &  \cellcolor{spatial}Viewpoint &  \cellcolor{spatial}Viewpoint &  \cellcolor{grey}Illumination &  \cellcolor{grey}Illumination &  \cellcolor{appearance}Temperature &  \cellcolor{appearance}Temperature &  \cellcolor{appearance}Exposure &  \cellcolor{appearance}Blur \\
    \midrule
    \multirow{5}{*}{\rotatebox{90}{Distance}} & \multirow{5}{*}{$\downarrow$}  & Random     &       51.22 &    39.09 &           32.36 &           50.83 &              33.37 &              45.99 &             14.21 &             41.07 &           50.14 &     22.41 \\
    {} & {} & Supervised &       27.50 &    35.40 &           34.66 &           43.93 &              29.72 &              40.57 &             10.18 &             33.03 &           25.55 &     25.87 \\
    {} & {} & Default    &       \underline{19.96} &    \underline{24.20} &           \underline{20.29} &           \textbf{39.96} &              \underline{18.53} &              \underline{37.48} &              \underline{6.60} &             \underline{17.34} &           \underline{17.96} &     \underline{19.15} \\
    {} & {} & \cellcolor{spatial_dark}Spatial    &       \cellcolor{spatial}\textbf{19.91} &    \cellcolor{spatial}\textbf{23.19} &           \cellcolor{spatial}\textbf{15.94} &           \cellcolor{spatial}\underline{41.80} &              \textbf{15.73} &              \textbf{37.28} &             11.31 &             54.28 &           34.45 &     32.52 \\
    {} & {} & \cellcolor{appearance_dark}Appearance     &       45.37 &    44.07 &           38.83 &           53.62 &              30.54 &              41.63 &              \cellcolor{appearance}\textbf{4.35} &              \cellcolor{appearance}\textbf{8.14} &           \cellcolor{appearance}\textbf{13.15} &     \cellcolor{appearance}\textbf{10.97} \\
    \midrule
    \multirow{5}{*}{\rotatebox{90}{Similarity}} & \multirow{5}{*}{$\uparrow$}  & Random     &        0.62 &     0.42 &            0.48 &            0.24 &               0.58 &               0.44 &              0.90 &              0.73 &            0.41 &      \underline{0.91} \\
    {} & {} & Supervised &        \underline{0.90} &     0.70 &            0.64 &            0.67 &               0.77 &               0.75 &              0.97 &              0.89 &            0.84 &      0.89 \\
    {} & {} & Default    &        \textbf{0.94} &     \underline{0.75} &            \underline{0.77} &            \underline{0.70} &               \underline{0.84} &               \underline{0.76} &              \underline{0.98} &              \underline{0.97} &            \underline{0.90} &      0.90 \\
    {} & {} & \cellcolor{spatial_dark}Spatial    &        \cellcolor{spatial}\textbf{0.94} &     \cellcolor{spatial}\textbf{0.84}$\bullet$ &            \cellcolor{spatial}\textbf{0.88}$\bullet$ &            \cellcolor{spatial}\textbf{0.72} &               \textbf{0.90}$\bullet$ &               \textbf{0.80}$\bullet$ &              0.94 &              0.78 &            0.69 &      0.82 \\
    {} & {} & \cellcolor{appearance_dark}Appearance     &        0.74 &     0.54 &            0.64 &            0.32 &               0.79 &               0.61 &              \cellcolor{appearance}\textbf{1.00} &              \cellcolor{appearance}\textbf{0.99} &            \cellcolor{appearance}\textbf{0.96}$\bullet$ &      \cellcolor{appearance}\textbf{0.98}$\bullet$ \\

    \bottomrule
    \end{tabular}
    }
\end{table*}

\keypoint{Results} The results in Table \ref{tab:realworld_invariances} group the benchmark datasets according to whether they exhibit spatial-like or appearance-like real-world translations. We note that the ALOI/ALOT benchmarks' illumination change condition moves a spotlight in a low ambient light scenario, effectively masking out different parts of the object. As it is unclear whether this corresponds to a appearance or spatial-like transformation, we color these separately. 
From the results we can see that our models have learned strong real-world invariances in many cases. For example, the spatial model has maximum similarity for all the (green) spatial-like transformations. The appearance model has maximum similarity for all the (orange) appearance-like transformations.
We carried out statistical hypothesis tests to determine which of the Default, Spatial, and Appearance models exhibit the most invariance to each type of transformation. These tests were carried out with a confidence level of 95\%, and statistically significant maxima are marked with a $\bullet$.
That the spatial model trained with crops and flips achieves stronger illumination invariance than one trained with colour augmentations on the ALOI/ALOT datasets is interesting, and highlights the importance of understanding the role of data augmentation better in representation learning.

\keypoint{Summary} In summary, we asked 
\emph{Q1: Whether invariances learned using data augmentation generalise to real-world transforms?} Grouping by appearance and spatial family invariances and real-world transforms, the answer is yes. To visualise this qualitatively, we report in Figure~\ref{fig:rank_correlation} of our supplementary materials the rank correlation (over our five models) between all pairs of synthetic and real-world invariances considered. From the results we can see a clear block structure that strong invariance to appearance-like synthetic transforms correlates with the appearance-like real-world transforms, and vice-versa. More formally, the statistical significance tests in Table~\ref{tab:realworld_invariances} show that five of eight comparisons show statistically significant impact of training augmentation with real-world transformation invariance.

For \emph{Q2: Is there a trade-off between learning different types of invariances?} A2: Yes. We have found that both for synthetic and real-world transforms, increasing appearance-style invariances decreases spatial-style ones and vice-versa. Models using the default set of augmentations suffer from this trade-off (we show that other state-of-the-art learners suffer similarly in Fig.~\ref{fig:v_d_vs_downstream_existing_models} of the supplement) Next, we look into whether a greater invariance to one family of transforms induces sensitivity in the other.

\cut{The only exception to this is that data with real-world illumination changes exhibits stronger correlation with spatial-style synthetic transforms.}

\begin{table*}[t]
    \centering
    \caption{Comparing models learned invariances on Causal3DIdent. Regression $R^2$ fit when predicting parameters from features. Our Appearance model is highly sensitive to spatial-style transforms and our Spatial model highly sensitive to appearance-style transforms. Our fused models exhibit strong predictive capability across the board.}
    \label{tab:causal3dident}
    \resizebox{1.0\linewidth}{!}{%
    \begin{tabular}{lcccccccccc}
    \toprule
    {} &  \cellcolor{appearance}Obj. Pos. x &  \cellcolor{appearance}Obj. Pos. y &  \cellcolor{appearance}Obj. Pos. z &  \cellcolor{appearance}Obj. Rot. x &  \cellcolor{appearance}Obj. Rot. y &  \cellcolor{appearance}Obj. Rot. z &  \cellcolor{grey}Spot. Pos. &  \cellcolor{spatial}Obj. Colour &  \cellcolor{spatial}Spot. Colour &  \cellcolor{spatial}Bkg. Colour \\
    \midrule
    \cellcolor{spatial_dark}Spatial     &         0.89 &         \underline{0.91} &         0.84 &         0.66 &         0.67 &         0.62 &         0.92 &         \cellcolor{spatial}\textbf{0.93} &           \cellcolor{spatial}\textbf{0.91} &         \cellcolor{spatial}\textbf{1.00} \\
    \cellcolor{appearance_dark}Appearance      &         \cellcolor{appearance}\underline{0.94} &         \cellcolor{appearance}\textbf{0.97} &         \cellcolor{appearance}\underline{0.91} &         \cellcolor{appearance}\underline{0.80} &         \cellcolor{appearance}\underline{0.84} &         \cellcolor{appearance}\underline{0.79} &         \underline{0.95} &         0.56 &           0.22 &         \underline{0.88} \\
    Spa+App     &         \underline{0.94} &         \textbf{0.97} &         \underline{0.91} &         \underline{0.80} &         0.82 &         0.77 &         \underline{0.95} &         \underline{0.92} &           \underline{0.90} &         \textbf{1.00} \\
    Def+Spa+App &         \textbf{0.95} &         \textbf{0.97} &         \textbf{0.94} &         \textbf{0.87} &         \textbf{0.89} &         \textbf{0.85} &         \textbf{0.96} &         \underline{0.92} &           0.89 &         \textbf{1.00} \\
    \bottomrule
    \end{tabular}
    }
\end{table*}

\subsection{Real-World Extrinsic Invariance Measurements}\label{sub:extrinsic}
To provide a different perspective on invariance to real-world transforms, we evaluate our features on the Causal3DIdent benchmark. In particular, we follow \cite{vonKugelgen2021Self-SupervisedStyle} in regressing real-world variables such as pose, object colour, light colour, etc.~from our features. A feature with complete spatial invariance would fail to predict pose, while one with colour invariance would fail to predict colour, etc.

\keypoint{Setup} We use kernel ridge regression 
with an RBF kernel, sample 20,480 training points and 40,960 test points and standardise images and targets. As the feature dimensionality of our models is much greater than of those in \cite{vonKugelgen2021Self-SupervisedStyle}, we expand the hyperparameter search space for $\alpha$ and $\gamma$ to $[10^{-4}, 10^{-3}, 10^{-2}, 10^{-1}, 10^{0}]$ and $[10^{-5}, 10^{-4}, 10^{-3}, 10^{-2}, 10^{-1}, 10^{0}, 10^{1}, 10^{2}]$, respectively.

\keypoint{Results} From the results in Table~\ref{tab:causal3dident}  this evaluation paradigm also confirms that learned invariances translate to some extent to real transformations. The appearance model obtains better performance on pose prediction tasks, while the spatial model obtains better performance on colour prediction tasks.

\section{Do Downstream Tasks Prefer Different Invariances?} \label{sec:downstream_tasks}
In the previous sections we have showed how contrastive training under data augmentation learns invariance to synthetic and real-world transformations. It also confirmed the colour/texture sensitivity of the Spatial model and the spatial sensitivity of the Appearance model. The Default model was found to always fall in between the two more specialised learners, with weaker invariance than one alternative but stronger than the other. 

In terms of real-world benchmarks, self-supervised methods are widely evaluated on ImageNet recognition, with the literature having a lesser focus and lack of consistency in evaluation of other non-recognition tasks. Since the default augmentations are largely chosen to optimise recognition benchmarks, there is a chance that it may be overfit to these tasks and perform less well on others. We therefore investigate how learned invariances affect a more diverse suite of real downstream tasks of interest, hypothesising that different features may be preferred, depending on the (in)variance needs of each downstream task. 

\begin{table}[t]
    \centering
    \caption{Downstream performances of our models. We report mean and standard deviation of 5-fold cross-validation on all data for each task. The differing performances in the tasks showcases how the Spatial and Appearance models capture important but \textit{different} properties necessary for wide transfer. Random refers to a randomly initialised feature extractor and `+' refers to feature concatenation. On the left datasets we report the classification accuracy and on the right the $R^2$ regression metric. Row colours indicate whether appearance or spatial turned out better for the given task.}
    \label{tab:downstream_tasks}
    \resizebox{\linewidth}{!}{%
    \begin{tabular}{l|cccccc|c}
    \toprule
    {} & \cellcolor{spatial}CIFAR10 & \cellcolor{spatial}Caltech101 & \cellcolor{spatial}Flowers & \cellcolor{appearance}300W & \cellcolor{appearance}CelebA & \cellcolor{appearance}LSPose &  Avg. \\
    \midrule
    Random             &  0.55 $\pm$ 0.004 &  0.25 $\pm$ 0.008 &  0.21 $\pm$ 0.009 &  0.24 $\pm$ 0.024 &  0.47 $\pm$ 0.002 &  0.10 $\pm$ 0.007 &  0.30 $\pm$ 0.009 \\
    Supervised         &  \textbf{0.98 $\pm$ 0.001} &  \textbf{0.90 $\pm$ 0.005} &  \textbf{0.86 $\pm$ 0.007} &  0.17 $\pm$ 0.028 &  0.49 $\pm$ 0.002 &  0.20 $\pm$ 0.015 &  0.60 $\pm$ 0.010 \\
    Default            &  0.96 $\pm$ 0.002 &  0.87 $\pm$ 0.006 &  0.83 $\pm$ 0.004 &  0.47 $\pm$ 0.014 &  0.60 $\pm$ 0.002 &  \textbf{0.29 $\pm$ 0.025} &  \textbf{0.67 $\pm$ 0.009} \\
    \cellcolor{spatial_dark}Spatial &  \cellcolor{spatial}0.92 $\pm$ 0.003 &  \cellcolor{spatial}0.65 $\pm$ 0.008 &  \cellcolor{spatial}0.74 $\pm$ 0.010 &  0.17 $\pm$ 0.030 &  0.49 $\pm$ 0.001 &  0.24 $\pm$ 0.020 &  0.54 $\pm$ 0.013 \\
    \cellcolor{appearance_dark}Appearance &  0.84 $\pm$ 0.003 &  0.57 $\pm$ 0.007 &  0.20 $\pm$ 0.009 &  \cellcolor{appearance}\textbf{0.68 $\pm$ 0.018} &  \cellcolor{appearance}\textbf{0.62 $\pm$ 0.003} &  \cellcolor{appearance}0.25 $\pm$ 0.021 &  0.53 $\pm$ 0.010 \\
    \midrule
    3$\times$Default          &  \textbf{0.96 $\pm$ 0.004} &  0.87 $\pm$ 0.004 &  \textbf{0.83 $\pm$ 0.007} &  0.52 $\pm$ 0.012 &  0.67 $\pm$ 0.001 &  0.31 $\pm$ 0.016 &  0.69 $\pm$ 0.007 \\
    Default($\times$3) &  \textbf{0.96 $\pm$ 0.002} &  \textbf{0.88 $\pm$ 0.003} &  0.82 $\pm$ 0.009 &  0.42 $\pm$ 0.030 &  0.65 $\pm$ 0.005 &  0.31 $\pm$ 0.028 &  0.67 $\pm$ 0.013 \\
    Spa+App            &  0.95 $\pm$ 0.002 &  0.74 $\pm$ 0.009 &  0.68 $\pm$ 0.005 &  0.62 $\pm$ 0.021 &  0.64 $\pm$ 0.002 &  0.29 $\pm$ 0.007 &  0.65 $\pm$ 0.008 \\
    Def+Spa+App        &  0.95 $\pm$ 0.003 &  0.86 $\pm$ 0.009 &  0.81 $\pm$ 0.006 &  \textbf{0.65 $\pm$ 0.020} &  \textbf{0.68 $\pm$ 0.002} &  \textbf{0.33 $\pm$ 0.010} &  \textbf{0.71 $\pm$ 0.008} \\
    \bottomrule
    \end{tabular}
    }
\end{table}




\keypoint{Experimental Details} Our suite of downstream tasks consists of object recognition on standard benchmarks \textbf{CIFAR10} \cite{Krizhevsky2009LearningImages}, \textbf{Caltech101} \cite{Fei-Fei2004LearningCategories} and \textbf{Flowers} \cite{Nilsback2008AutomatedClasses}; as well as a set of spatially sensitive tasks including facial landmark detection on \textbf{300W} \cite{Sagonas2016300Challenge} and \textbf{CelebA} \cite{liu2015faceattributes}, and pose estimation on \textbf{Leeds Sports Pose}~\cite{Johnson2010ClusteredEstimation}. We freeze the backbones and extract features from just after the average pooling layer of the ResNet50 architectures. We fit a ridge or logistic regression model on these features, depending on the task in question. To tune the $\ell2$ regularisation value we perform 5-fold cross-validation over a grid of 45 logarithmically spaced values between $10^{-6}$ to $10^5$, following \cite{Ericsson2021HowTransfer,Chen2020ARepresentations}. We report the mean and standard deviation for the hyperparameter choice with highest mean. The performance is reported as accuracies (between 0 and 1) for classification tasks and $R^2$ values for regression tasks. For comparison we also evaluate random and supervised backbones.

\keypoint{Results} \Cref{tab:downstream_tasks} shows the linear readout performance on all tasks considered. On the datasets most similar to ImageNet: CIFAR10, Caltech101 and Flowers, the Default or Supervised models achieve the highest classification accuracy, followed by the Spatial and then the Appearance model. On the spatially sensitive tasks the Appearance model outperforms the Spatial model substantively, with the Appearance model performing best overall on 300W. These results show some evidence that the Default (and to a lesser extent Spatial model) model is well suited for object recognition on ImageNet-like datasets, but both are weak in comparison to a model with more spatial sensitivity when solving the pose-related tasks. Overall this supports the hypothesis that different (in)variances are required for best performance on different types of tasks. To answer \emph{Q3: Do different downstream tasks of interest benefit from different invariances?} A3: Yes. On our classification tasks, representations trained on default or spatial-style augmentations dominate. Pose-related tasks benefit from appearance-style augmentations, where default augmentations under-perform.



\keypoint{Improving Performance Through Feature Fusion} Our previous analyses show that different real-world tasks prefer different invariances. The Default model tries to satisfy them all by using a mix of augmentations to obtain a moderate amount of invariance to all transformations (Sec~\ref{sec:synthetic-invariance}), but appearance/spatial specialised features can be better for particular tasks (Table~\ref{tab:downstream_tasks}, top). We therefore explore whether a fusion of specialised features can perform competitively across the board. In particular we explore Spatial-Appearance (Spa+App) fusion, as well as three way Default-Spatial-Appearance (Def+Spa+App) fusion.

\keypoint{Experimental Details} The evaluation follows the setup described above, but as our fused representations have higher dimensionality, we shift the $\ell2$ search space for Spa+App to $10^{-5}$ to $10^6$ and Def+Spa+App $10^{-4}$ to $10^7$. Finally, to compare the concatenated features of Def+Spa+App, we evaluate a second Default model with a $3 \times$ wider architecture -- ResNet50($\times 3$) -- which was trained with MoCo-v2 for 200 epochs on ImageNet like our other models and uses the same hyperparameter search space as Def+Spa+App. A final baseline consisting of three fused separately trained Default models forms $3\times$Default.

\keypoint{Results} From the results in Table~\ref{tab:downstream_tasks} (bottom) and Table~\ref{tab:causal3dident} (bottom), we can see that the Spa+App model and Def+Spa+App fusion models perform strongly across the board. While the $3\times$Default and Default($\times 3$) models are unsurprisingly best for recognition tasks, this is only by a small margin; while the 3-way Def+Spa+App fusion is dramatically better for 300W, and the most consistent performer across the board. To answer \emph{Q4: Is there a simple way to achieve high performance across all tasks?} A4: Yes. We fuse multiple representations tuned for different (in)variances for consistently strong performance across all tasks considered.

This result is noteworthy, as a goal of self-supervised learning is to provide a single feature that provides excellent performance for diverse downstream tasks. While we showed the default model falls down in this regard, our fused feature provides reliable performance across the board. We therefore recommend it to practitioners who want a single feature with which to perform diverse tasks.

\section{Discussion} We have performed the first thorough evaluation of self-supervised learning in terms of augmentations used for training, and resulting downstream invariance and task impact. In particular we showed that: (1) CNNs trained contrastively do learn invariances corresponding to the augmentations used and specialising CNNs to particular appearance/spatial augmentations can lead to greater corresponding invariances (Table~\ref{tab:synthetic_invariances}). Furthermore, learning invariances to synthetic transforms does provide a degree of invariance to corresponding real-world transforms (Table~\ref{tab:realworld_invariances}, Fig~\ref{fig:rank_correlation}).
(2) Different real-world downstream tasks do prefer features providing different invariances (Table~\ref{tab:downstream_tasks}, Fig.~\ref{fig:v_d_vs_downstream_existing_models}), and invariance-specialised features can sometimes outperform the standard default augmentation, e.g., for spatially sensitive tasks. (3) Fusing features tuned for different types of invariances provides a consistently high performing strategy (Table~\ref{tab:downstream_tasks}). This outperforms the default model on pose related tasks, suggesting that it was over-tuned for recognition. Our feature ensemble strategy is promising for providing high performance \emph{general purpose} real-world features. 
Based on these results we encourage the SSL community to evaluate on more diverse downstream task types.


\subsection*{Acknowledgements}
This work was supported by the Engineering and Physical Sciences Research Council (EPSRC) Grant number EP/S000631/1; and the MOD University Defence Research Collaboration (UDRC) in Signal Processing. This project was supported by the Royal Academy of Engineering under the Research Fellowship programme.

%
%
\bibliography{references,extraref}

\begin{thebibliography}{51}
\providecommand{\natexlab}[1]{#1}
\providecommand{\url}[1]{\texttt{#1}}
\expandafter\ifx\csname urlstyle\endcsname\relax
  \providecommand{\doi}[1]{doi: #1}\else
  \providecommand{\doi}{doi: \begingroup \urlstyle{rm}\Url}\fi

\bibitem[Afifi et~al.(2021)Afifi, Derpanis, Ommer, and
  Brown]{Afifi2021LearningCorrection}
Mahmoud Afifi, Konstantinos~G. Derpanis, Björn Ommer, and Michael~S. Brown.
\newblock {Learning Multi-Scale Photo Exposure Correction}.
\newblock In \emph{CVPR}, 2021.

\bibitem[Bakhtiari et~al.(2021)Bakhtiari, Mineault, Lillicrap, Pack, and
  Richards]{Bakhtiari2021TheLearning}
Shahab Bakhtiari, Patrick Mineault, Tim Lillicrap, Christopher~C. Pack, and
  Blake~A. Richards.
\newblock {The functional specialization of visual cortex emerges from training
  parallel pathways with self-supervised predictive learning}.
\newblock In \emph{NeurIPS}, 2021.

\bibitem[Biscione and Bowers(2020)]{Biscione2020LearningCNNs}
Valerio Biscione and Jeffrey Bowers.
\newblock {Learning Translation Invariance in CNNs}.
\newblock In \emph{2nd Workshop on Shared Visual Representations in Human and
  Machine Intelligence, NeurIPS}, 2020.

\bibitem[Burghouts and
  Geusebroek(2009)]{Burghouts2009Material-specificFeatures}
Gertjan~J Burghouts and Jan-Mark Geusebroek.
\newblock {Material-specific adaptation of color invariant features}.
\newblock \emph{Pattern Recognition Letters}, 2009.

\bibitem[Caron et~al.(2020)Caron, Misra, Mairal, Goyal, Bojanowski, and
  Joulin]{Caron2020UnsupervisedAssignments}
Mathilde Caron, Ishan Misra, Julien Mairal, Priya Goyal, Piotr Bojanowski, and
  Armand Joulin.
\newblock {Unsupervised Learning of Visual Features by Contrasting Cluster
  Assignments}.
\newblock In \emph{NeurIPS}, 2020.

\bibitem[Caron et~al.(2021)Caron, Touvron, Misra, J\'egou, Mairal, Bojanowski,
  and Joulin]{caron2021emerging}
Mathilde Caron, Hugo Touvron, Ishan Misra, Herv\'e J\'egou, Julien Mairal,
  Piotr Bojanowski, and Armand Joulin.
\newblock Emerging properties in self-supervised vision transformers.
\newblock In \emph{ICCV}, 2021.

\bibitem[Chen et~al.(2020{\natexlab{a}})Chen, Kornblith, Norouzi, and
  Hinton]{Chen2020ARepresentations}
Ting Chen, Simon Kornblith, Mohammad Norouzi, and Geoffrey Hinton.
\newblock {A Simple Framework for Contrastive Learning of Visual
  Representations}.
\newblock In \emph{ICML}, 2020{\natexlab{a}}.

\bibitem[Chen et~al.(2020{\natexlab{b}})Chen, Kornblith, Swersky, Norouzi, and
  Hinton]{Chen2020BigLearners}
Ting Chen, Simon Kornblith, Kevin Swersky, Mohammad Norouzi, and Geoffrey
  Hinton.
\newblock {Big Self-Supervised Models are Strong Semi-Supervised Learners}.
\newblock In \emph{NeurIPS}, 2020{\natexlab{b}}.

\bibitem[Chen and He(2021)]{Chen2021ExploringLearning}
Xinlei Chen and Kaiming He.
\newblock {Exploring Simple Siamese Representation Learning}.
\newblock In \emph{CVPR}, 2021.

\bibitem[Chen et~al.(2020{\natexlab{c}})Chen, Fan, Girshick, and
  He]{Chen2020ImprovedLearning}
Xinlei Chen, Haoqi Fan, Ross Girshick, and Kaiming He.
\newblock {Improved Baselines with Momentum Contrastive Learning}.
\newblock \emph{arXiv}, 2020{\natexlab{c}}.

\bibitem[Cree et~al.(2016)Cree, Perrone, Anthonys, Garnett, and
  Gouk]{cree2016estimating}
Michael~J Cree, John~A Perrone, Gehan Anthonys, Aden~C Garnett, and Henry Gouk.
\newblock Estimating heading direction from monocular video sequences using
  biologically-based sensors.
\newblock In \emph{IVCNZ}, 2016.

\bibitem[Cubuk et~al.(2019)Cubuk, Zoph, Mane, Vasudevan, and
  Le]{Cubuk2019AutoAugment:Data}
Ekin~D. Cubuk, Barret Zoph, Dandelion Mane, Vijay Vasudevan, and Quoc~V. Le.
\newblock {AutoAugment: Learning Augmentation Policies from Data}.
\newblock In \emph{CVPR}, 2019.

\bibitem[Cubuk et~al.(2020)Cubuk, Zoph, Shlens, and
  Le]{Cubuk2020RandAugment:Space}
Ekin~D. Cubuk, Barret Zoph, Jonathon Shlens, and Quoc~V. Le.
\newblock {RandAugment: Practical automated data augmentation with a reduced
  search space}.
\newblock In \emph{NeurIPS}, 2020.

\bibitem[Deng et~al.(2009)Deng, Dong, Socher, Li, Li, and
  Fei-Fei]{Deng2009ImageNet:Database}
J~Deng, W~Dong, R~Socher, L.-J. Li, K~Li, and L~Fei-Fei.
\newblock {ImageNet: A Large-Scale Hierarchical Image Database}.
\newblock In \emph{CVPR}, 2009.

\bibitem[Ebrahimpour et~al.(2019)Ebrahimpour, Li, Yu, Reese, Moghtaderi, Yang,
  and Noelle]{Ebrahimpour2019Ventral-DorsalAttention}
Mohammad~K. Ebrahimpour, Jiayun Li, Yen-Yun Yu, Jackson~L. Reese, Azadeh
  Moghtaderi, Ming-Hsuan Yang, and David~C. Noelle.
\newblock {Ventral-Dorsal Neural Networks: Object Detection via Selective
  Attention}.
\newblock In \emph{WACV}, 2019.

\bibitem[Ericsson et~al.(2021)Ericsson, Gouk, and
  Hospedales]{Ericsson2021HowTransfer}
Linus Ericsson, Henry Gouk, and Timothy~M. Hospedales.
\newblock {How Well Do Self-Supervised Models Transfer?}
\newblock In \emph{CVPR}, 2021.

\bibitem[Ericsson et~al.(2022)Ericsson, Gouk, Loy, and
  Hospedales]{ericsson2021self}
Linus Ericsson, Henry Gouk, Chen~Change Loy, and Timothy~M Hospedales.
\newblock Self-supervised representation learning: Introduction, advances and
  challenges.
\newblock \emph{IEEE Signal Processing Magazine}, 2022.

\bibitem[Fei-Fei et~al.(2004)Fei-Fei, Fergus, and
  Perona]{Fei-Fei2004LearningCategories}
Li~Fei-Fei, Rob Fergus, and Pietro Perona.
\newblock {Learning generative visual models from few training examples: An
  incremental bayesian approach tested on 101 object categories}.
\newblock In \emph{CVPR Workshops}, 2004.

\bibitem[Geusebroek et~al.(2005)Geusebroek, Burghouts, and
  Smeulders]{Geusebroek2005TheImages}
Jan-Mark Geusebroek, Gertjan~J. Burghouts, and Arnold~W.M. Smeulders.
\newblock {The Amsterdam Library of Object Images}.
\newblock \emph{International Journal of Computer Vision}, 2005.

\bibitem[Goodale and Milner(1992)]{Goodale1992SeparateAction}
Melvyn~A. Goodale and A.~David Milner.
\newblock {Separate visual pathways for perception and action}.
\newblock \emph{Trends in Neurosciences}, 1992.

\bibitem[Goodfellow et~al.(2009)Goodfellow, Le, Saxe, Lee, and
  Ng]{Goodfellow2009MeasuringNetworks}
Ian~J Goodfellow, Quoc~V Le, Andrew~M Saxe, Honglak Lee, and Andrew~Y Ng.
\newblock {Measuring Invariances in Deep Networks}.
\newblock In \emph{NeurIPS}, 2009.

\bibitem[Grill et~al.(2020)Grill, Strub, Altch{\'{e}}, Tallec, Richemond,
  Buchatskaya, Doersch, Pires, Guo, Azar, Piot, Kavukcuoglu, Munos, and
  Valko]{Grill2020BootstrapLearning}
Jean-Bastien Grill, Florian Strub, Florent Altch{\'{e}}, Corentin Tallec,
  Pierre~H. Richemond, Elena Buchatskaya, Carl Doersch, Bernardo~Avila Pires,
  Zhaohan~Daniel Guo, Mohammad~Gheshlaghi Azar, Bilal Piot, Koray Kavukcuoglu,
  Rémi Munos, and Michal Valko.
\newblock {Bootstrap Your Own Latent: A New Approach to Self-Supervised
  Learning}.
\newblock In \emph{NeurIPS}, 2020.

\bibitem[He et~al.(2016)He, Zhang, Ren, and Sun]{He2016DeepRecognition}
Kaiming He, Xiangyu Zhang, Shaoqing Ren, and Jian Sun.
\newblock {Deep residual learning for image recognition}.
\newblock In \emph{CVPR}, 2016.

\bibitem[He et~al.(2019)He, Fan, Wu, Xie, and Girshick]{He2019MomentumLearning}
Kaiming He, Haoqi Fan, Yuxin Wu, Saining Xie, and Ross Girshick.
\newblock {Momentum Contrast for Unsupervised Visual Representation Learning}.
\newblock In \emph{CVPR}, 2019.

\bibitem[Jang et~al.(2017)Jang, Vijayanarasimhan, Pastor, Ibarz, and
  Levine]{Jang2017End-to-EndGrasping}
Eric Jang, Sudheendra Vijayanarasimhan, Peter Pastor, Julian Ibarz, and Sergey
  Levine.
\newblock {End-to-End Learning of Semantic Grasping}.
\newblock In \emph{CoRL}, 2017.

\bibitem[Jing and Tian(2020)]{Jing2020Self-supervisedSurvey}
Longlong Jing and Yingli Tian.
\newblock {Self-supervised Visual Feature Learning with Deep Neural Networks: A
  Survey}.
\newblock \emph{IEEE Transactions on Pattern Analysis and Machine
  Intelligence}, 2020.

\bibitem[Johnson and Everingham(2010)]{Johnson2010ClusteredEstimation}
Sam Johnson and Mark Everingham.
\newblock {Clustered Pose and Nonlinear Appearance Models for Human Pose
  Estimation}.
\newblock In \emph{BMVC}, 2010.

\bibitem[Kayhan and van Gemert(2020)]{Kayhan2020OnLocation}
Osman~Semih Kayhan and Jan~C. van Gemert.
\newblock {On Translation Invariance in CNNs: Convolutional Layers can Exploit
  Absolute Spatial Location}.
\newblock In \emph{CVPR}, 2020.

\bibitem[Krizhevsky and Hinton(2009)]{Krizhevsky2009LearningImages}
Alex Krizhevsky and Geoffrey Hinton.
\newblock {Learning Multiple Layers of Features from Tiny Images}.
\newblock \emph{arXiv}, 2009.

\bibitem[Kruger et~al.(2013)Kruger, Janssen, Kalkan, Lappe, Leonardis, Piater,
  Rodriguez-Sanchez, and Wiskott]{kruger2013deepHierarchies}
N.~Kruger, P.~Janssen, S.~Kalkan, M.~Lappe, A.~Leonardis, J.~Piater, A.~J.
  Rodriguez-Sanchez, and L.~Wiskott.
\newblock Deep hierarchies in the primate visual cortex: What can we learn for
  computer vision?
\newblock \emph{IEEE Transactions on Pattern Analysis and Machine
  Intelligence}, 2013.

\bibitem[Li et~al.(2021)Li, Zhou, Xiong, Socher, and
  Hoi]{Li2021PrototypicalRepresentations}
Junnan Li, Pan Zhou, Caiming Xiong, Richard Socher, and Steven C.~H. Hoi.
\newblock {Prototypical Contrastive Learning of Unsupervised Representations}.
\newblock In \emph{ICLR}, 2021.

\bibitem[Lin et~al.(2014)Lin, Maire, Belongie, Hays, Perona, Ramanan,
  Doll{\'{a}}r, and Zitnick]{Lin2014MicrosoftContext}
Tsung~Yi Lin, Michael Maire, Serge Belongie, James Hays, Pietro Perona, Deva
  Ramanan, Piotr Doll{\'{a}}r, and C.~Lawrence Zitnick.
\newblock {Microsoft COCO: Common objects in context}.
\newblock In \emph{ECCV}, 2014.

\bibitem[Liu et~al.(2015)Liu, Luo, Wang, and Tang]{liu2015faceattributes}
Ziwei Liu, Ping Luo, Xiaogang Wang, and Xiaoou Tang.
\newblock Deep learning face attributes in the wild.
\newblock In \emph{ICCV}, 2015.

\bibitem[Mikolajczyk and Schmid(2005)]{mikolajczyk2005descriptor_eval}
K.~Mikolajczyk and C.~Schmid.
\newblock A performance evaluation of local descriptors.
\newblock \emph{IEEE Transactions on Pattern Analysis and Machine
  Intelligence}, 2005.

\bibitem[Nene et~al.(1996)Nene, Nayar, and Murase]{Nene1996ColumbiaCOIL100}
S.~A. Nene, S.~K. Nayar, and H.~Murase.
\newblock {Columbia Object Image Library (COIL100)}.
\newblock Technical report, 1996.

\bibitem[Nilsback and Zisserman(2008)]{Nilsback2008AutomatedClasses}
Maria~Elena Nilsback and Andrew Zisserman.
\newblock {Automated flower classification over a large number of classes}.
\newblock In \emph{ICVGIP}, 2008.

\bibitem[Oord et~al.(2017)Oord, Vinyals, and
  Kavukcuoglu]{Oord2017NeuralLearning}
Aaron van~den Oord, Oriol Vinyals, and Koray Kavukcuoglu.
\newblock {Neural Discrete Representation Learning}.
\newblock \emph{NeurIPS}, 2017.

\bibitem[Paszke et~al.(2019)Paszke, Gross, Massa, Lerer, Bradbury, Chanan,
  Killeen, Lin, Gimelshein, Antiga, Desmaison, Kopf, Yang, DeVito, Raison,
  Tejani, Chilamkurthy, Steiner, Fang, Bai, and
  Chintala]{Paszke2019PyTorch:Library}
Adam Paszke, Sam Gross, Francisco Massa, Adam Lerer, James Bradbury, Gregory
  Chanan, Trevor Killeen, Zeming Lin, Natalia Gimelshein, Luca Antiga, Alban
  Desmaison, Andreas Kopf, Edward Yang, Zachary DeVito, Martin Raison, Alykhan
  Tejani, Sasank Chilamkurthy, Benoit Steiner, Lu~Fang, Junjie Bai, and Soumith
  Chintala.
\newblock {PyTorch: An Imperative Style, High-Performance Deep Learning
  Library}.
\newblock In \emph{NeurIPS}, 2019.

\bibitem[Perrone et~al.(2019)Perrone, Cree, and Hedayati]{perrone2019using}
John~A. Perrone, Michael~J. Cree, and Mohammad Hedayati.
\newblock Using the properties of primate motion sensitive neurons to extract
  camera motion and depth from brief 2-d monocular image sequences.
\newblock In \emph{CAIP}, 2019.

\bibitem[Purushwalkam and Gupta(2020)]{Purushwalkam2020DemystifyingBiases}
Senthil Purushwalkam and Abhinav Gupta.
\newblock {Demystifying Contrastive Self-Supervised Learning: Invariances,
  Augmentations and Dataset Biases}.
\newblock In \emph{NeurIPS}, 2020.

\bibitem[Rim et~al.(2020)Rim, Lee, Won, and Cho]{Rim2020Real-WorldAlgorithms}
Jaesung Rim, Haeyun Lee, Jucheol Won, and Sunghyun Cho.
\newblock {Real-World Blur Dataset for Learning and Benchmarking Deblurring
  Algorithms}.
\newblock In \emph{ECCV}, 2020.

\bibitem[Sagonas et~al.(2016)Sagonas, Antonakos, Tzimiropoulos, Zafeiriou, and
  Pantic]{Sagonas2016300Challenge}
Christos Sagonas, Epameinondas Antonakos, Georgios Tzimiropoulos, Stefanos
  Zafeiriou, and Maja Pantic.
\newblock {300 Faces In-The-Wild Challenge}.
\newblock \emph{Image and Vision Computing}, 2016.

\bibitem[Scharstein et~al.(2014)Scharstein, Hirschm{\"{u}}ller, Kitajima,
  Krathwohl, Ne{\v{s}}i{\'{c}}, Wang, and
  Westling]{Scharstein2014High-ResolutionTruth}
Daniel Scharstein, Heiko Hirschm{\"{u}}ller, York Kitajima, Greg Krathwohl,
  Nera Ne{\v{s}}i{\'{c}}, Xi~Wang, and Porter Westling.
\newblock {High-Resolution Stereo Datasets with Subpixel-Accurate Ground
  Truth}.
\newblock In \emph{GCPR}, 2014.

\bibitem[Shorten and Khoshgoftaar(2019)]{Shorten2019ALearning}
Connor Shorten and Taghi~M. Khoshgoftaar.
\newblock {A survey on Image Data Augmentation for Deep Learning}.
\newblock \emph{Journal of Big Data}, 2019.

\bibitem[Van~Horn et~al.(2021)Van~Horn, Cole, Beery, Wilber, Belongie, and
  Mac~Aodha]{VanHorn2021BenchmarkingCollections}
Grant Van~Horn, Elijah Cole, Sara Beery, Kimberly Wilber, Serge Belongie, and
  Oisin Mac~Aodha.
\newblock {Benchmarking Representation Learning for Natural World Image
  Collections}.
\newblock In \emph{CVPR}, 2021.

\bibitem[von K{\"{u}}gelgen et~al.(2021)von K{\"{u}}gelgen, Sharma, Gresele,
  Brendel, Sch{\"{o}}lkopf, Besserve, and
  Locatello]{vonKugelgen2021Self-SupervisedStyle}
Julius von K{\"{u}}gelgen, Yash Sharma, Luigi Gresele, Wieland Brendel,
  Bernhard Sch{\"{o}}lkopf, Michel Besserve, and Francesco Locatello.
\newblock {Self-Supervised Learning with Data Augmentations Provably Isolates
  Content from Style}.
\newblock In \emph{NeurIPS}, 2021.

\bibitem[Wang and Isola(2020)]{Wang2020UnderstandingHypersphere}
Tongzhou Wang and Phillip Isola.
\newblock {Understanding Contrastive Representation Learning through Alignment
  and Uniformity on the Hypersphere}.
\newblock In \emph{ICML}, 2020.

\bibitem[Wang et~al.(2022)Wang, Zhang, Wang, Yang, and
  Lin]{Wang2022ChaosLearning}
Yifei Wang, Qi~Zhang, Yisen Wang, Jiansheng Yang, and Zhouchen Lin.
\newblock {Chaos is a Ladder: A New Understanding of Contrastive Learning}.
\newblock In \emph{ICLR}, 2022.

\bibitem[Wen and Li(2021)]{Wen2021TowardLearning}
Zixin Wen and Yuanzhi Li.
\newblock {Toward Understanding the Feature Learning Process of Self-supervised
  Contrastive Learning}.
\newblock In \emph{ICML}, 2021.

\bibitem[Xiao et~al.(2021)Xiao, Wang, Efros, and Darrell]{Xiao2021WhatLearning}
Tete Xiao, Xiaolong Wang, Alexei~A. Efros, and Trevor Darrell.
\newblock {What Should Not Be Contrastive in Contrastive Learning}.
\newblock In \emph{ICLR}, 2021.

\bibitem[Zbontar et~al.(2021)Zbontar, Jing, Misra, LeCun, and
  Deny]{Zbontar2021BarlowReduction}
Jure Zbontar, Li~Jing, Ishan Misra, Yann LeCun, and Stéphane Deny.
\newblock {Barlow Twins: Self-Supervised Learning via Redundancy Reduction}.
\newblock In \emph{ICML}, 2021.

\end{thebibliography}

\clearpage

\appendix
\counterwithin{figure}{section}
\counterwithin{table}{section}
\counterwithin{lstlisting}{section}
\counterwithin{equation}{section}

\section{Data augmentation for pre-training}

We use the strong contrastive baseline MoCo-v2 \cite{Chen2020ImprovedLearning} for pre-training our models and use its augmentation policy as our basis for our experiments. Our Default model is trained using the full set of augmentations detailed below in PyTorch \cite{Paszke2019PyTorch:Library} code.
\begin{lstlisting}[language=Python, caption=Default augmentation policy]
transforms.Compose([
    transforms.RandomResizedCrop(224, scale=(0.2, 1.)),
    transforms.RandomApply([
        transforms.ColorJitter(0.4, 0.4, 0.4, 0.1)
    ], p=0.8),
    transforms.RandomGrayscale(p=0.2),
    transforms.RandomApply([moco.loader.GaussianBlur([.1, 2.])], p=0.5),
    transforms.RandomHorizontalFlip(),
    transforms.ToTensor(),
    transforms.Normalize(mean=[0.485, 0.456, 0.406], std=[0.229, 0.224, 0.225])
])
\end{lstlisting}

The Spatial model only uses resized crops and horizontal flips.
\begin{lstlisting}[language=Python, caption=Spatial augmentation policy]
transforms.Compose([
    transforms.RandomResizedCrop(224, scale=(0.2, 1.)),
    transforms.RandomHorizontalFlip(),
    transforms.ToTensor(),
    transforms.Normalize(mean=[0.485, 0.456, 0.406], std=[0.229, 0.224, 0.225])
])
\end{lstlisting}

And finally the Appearance model uses grayscale, color jitter and blurring.
\begin{lstlisting}[language=Python, caption=Appearance augmentation policy]
transforms.Compose([
    transforms.Resize(224),
    transforms.CenterCrop(224),
    transforms.RandomApply([
        transforms.ColorJitter(0.4, 0.4, 0.4, 0.1)
    ], p=0.8),
    transforms.RandomGrayscale(p=0.2),
    transforms.RandomApply([moco.loader.GaussianBlur([.1, 2.])], p=0.5),
    transforms.ToTensor(),
    transforms.Normalize(mean=[0.485, 0.456, 0.406], std=[0.229, 0.224, 0.225])
])
\end{lstlisting}

The Random model was not pre-trained and thus did not use any augmentations. The Supervised model was pre-trained using both spatial and appearance-based augmentation (crop, h.~flip and a version of color jitter) \cite{He2016DeepRecognition}.

\section{Synthetic transforms}

For Table~\ref{tab:synthetic_invariances} we compute the invariance metrics based on the following synthetic transforms.
\begin{itemize}
    \item Resized crop: We generate 256 crops with anchor points positioned between 0 and 64 pixels from the top and left of the original image (of size $256 \times 256$). A crop is between 25\% and 75\% of the image in both height and width. After, the crop is resized to $224 \times 224$.
    \item Horizontal flip: A single horizontally flipped image is generated in addition to the unaugmented image.
    \item Vertical flip: A single vertically flipped image is generated in addition to the unaugmented image.
    \item Scale: We generate 256 images rescaled between $\frac{1}{4}$ to $2$ times its original size.
    \item Shear: We generate 256 images with horizontal and vertical shear of -160 to 160 degrees.
    \item Rotation: We generate 256 images with rotation angles between 0 and 360 degrees.
    \item Translation: We generate 256 images with horizontal and vertical translation of -16 to 16 pixels.
    \item Deform: We generate 256 images with the ElasticTransform function of the albumentations package, with $\sigma$ between 10 and 50.
    \item Grayscale: A single grayscale image is generated in addition to the unaugmented image.
    \item Brightness: We generate 256 images where brightness is $\frac{1}{4}$ to $5$ times its original value.
    \item Contrast: We generate 256 images where contrast is $\frac{1}{4}$ to $5$ times its original value.
    \item Saturation: We generate 256 images where saturation is $\frac{1}{4}$ to $5$ times its original value.
    \item Hue: We generate 256 images where hue is set to one of 5 values spread over the colour circle.
    \item Blur: We generate 256 images with Gaussian blur where $\sigma$ is between $10^{-5}$ to $20$.
    \item Sharpness: We generate 256 images where the sharpness is adjusted by a factor of $1$ to $30$.
    \item Equalize: A single image with an equalized histogram is generated in addition to the unaugmented image.
    \item Posterize: We generate seven images by reducing the number of bits for each colour channel to 1-7 in addition to the full 8-bit unaugmented image.
    \item Invert: A single image with inverted colours is generated in addition to the unaugmented image.
\end{itemize}

\section{Invariance measurement details and further results}
\subsection{Measuring invariances} \label{sec:measure}

A key contribution of this paper is measuring the degree of invariance to various synthetic and real-world transformations. Previous studies have focused on measuring invariance at the neuronal level~\cite{Goodfellow2009MeasuringNetworks}. We consider instead the invariance properties of entire feature vectors under input transformations. To this end we explore two metrics.

\keypoint{Mahalanobis distance}
A vector can be said to be invariant to a transformation if it remains unchanged after applying that transformation. We can measure the invariances of a feature extractor model by looking at how much its feature vectors change under different transformations. Given a pre-trained feature extractor $f$, whose feature space has a covariance of $\bm{\Sigma}$, a transformation $t_\phi$ parameterised by $\phi$ and an image $x$, we compute the variance of $f$ to transformation $t_\phi$ as the Mahalanobis distance
\begin{align}
    l_{f}^{t_\phi}(x) &= \sqrt{\Big(f(x) - f(t_\phi(x))\Big) \bm{\Sigma}^{-1} \Big(f(x) - f(t_\phi(x))\Big)^T} = \|Gf(x) - Gf(t_\phi(x))\|_2
\end{align}
where $\bm{\Sigma}^{-1} = GG^T$, and $G$ can be computed using the Cholesky decomposition.

\keypoint{Cosine similarity}
Alternatively, we can measure the angle instead of the distance by first standardising the vectors using the mean feature, $\bar{f}$, of $f$ and $G$, giving us
\begin{gather}
    z = G \Big(\bar{f} - f(x)\Big), \qquad z_{t_\phi(x)} = G \Big(\bar{f} - f(t_\phi(x))\Big),
\end{gather}
and then using cosine similarity to measure the angle between features, giving us an invariance measure of
\begin{equation}
    l_{f}^{t_\phi}(x) = \frac{z \cdot z_{t_\phi(x)}}{\|z\| \, \|z_{t_\phi(x)}\|}.
\end{equation}


The distance or similarity is computed over a range of transformation parameters, $\phi \in \Phi$---e.g.~from $0\degree$ to $360\degree$ for rotation. Additionally, we average over all images in a dataset $D$. The global measurement is then

\begin{equation}
    L_{f}^{T_\Phi}(D) = \frac{1}{|D| |\Phi|} \sum_{x \in D} \sum_{\phi \in \Phi} l_{f}^{t_\phi}(x),
\end{equation}

where $T_\Phi = \{t_\phi\}_{\phi \in \Phi}$. A model with zero Mahalanobis distance (variance) to a transformation is invariant to it. Likewise, a model with maximum cosine similarity is invariant.

\begin{table}[t]
    \centering
    \caption{Top left: alignment metric on synthetic transforms, and top right: uniformity on the three augmentation families. Bottom left: alignment metric on real-world transforms and, bottom right: average confusion ratio (ACR) on the three augmentation families ($C$=10, $k$=10).}
    \label{tab:alignment}
    \resizebox{1.\linewidth}{!}{%
    \begin{tabular}{l|cccccccccccccccccc} 
    \toprule
    {} & \multicolumn{18}{c}{Alignment} \\ 
    \midrule
    {} &  \cellcolor{spatial}Crop &  \cellcolor{spatial}H flip &  \cellcolor{spatial}V flip &  \cellcolor{spatial}Scale &  \cellcolor{spatial}Shear &  \cellcolor{spatial}Rotat. &  \cellcolor{spatial}Transl. &  \cellcolor{spatial}Deform &  \cellcolor{appearance}Graysc. &  \cellcolor{appearance}Bright. &  \cellcolor{appearance}Contr. &  \cellcolor{appearance}Satur. &  \cellcolor{appearance}Hue &  \cellcolor{appearance}Blur &  \cellcolor{appearance}Sharpn. &  \cellcolor{appearance}Equal. &  \cellcolor{appearance}Poster. &  \cellcolor{appearance}Invert \\ 
    \midrule
    Random     &          1.94 &    0.88 &    0.93 &   1.66 &   1.92 &      1.86 &         1.32 &    1.61 &       0.38 &        1.66 &      0.97 &        0.81 & 0.80 &     1.04 &       0.97 &      0.61 &       0.56 &    0.98 \\ 
    Supervised &          1.63 &    0.16 &    0.58 &   0.86 &   1.77 &      1.14 &         0.27 &    0.61 &       0.38 &        0.92 &      0.72 &        0.43 & 0.72 &     1.42 &       0.91 &      0.31 &       0.45 &    0.72 \\ 
    Default    &          1.63 &    0.11 &    0.53 &   0.74 &   1.77 &      1.09 &         0.15 &    0.50 &       0.09 &        0.60 &      0.34 &        0.12 & 0.14 &     1.38 &       0.60 &      0.20 &       0.34 &    \textbf{0.67} \\ 
    \cellcolor{spatial_dark}Spatial    &          \cellcolor{spatial}\textbf{1.50} &    \cellcolor{spatial}\textbf{0.07} &    \cellcolor{spatial}\textbf{0.24} &   \cellcolor{spatial}\textbf{0.58} &   \cellcolor{spatial}\textbf{1.48} &      \cellcolor{spatial}\textbf{0.61} &         \cellcolor{spatial}\textbf{0.10} &    \cellcolor{spatial}\textbf{0.36} &       0.78 &        1.28 &      1.28 &        0.91 & 1.27 &     1.52 &       1.06 &      0.62 &       0.76 &    0.82 \\ 
    \cellcolor{appearance_dark}Appearance     &          1.95 &    0.75 &    0.82 &   1.48 &   1.94 &      1.81 &         0.58 &    1.30 &       \cellcolor{appearance}\textbf{0.01} &        \cellcolor{appearance}\textbf{0.33} &      \cellcolor{appearance}\textbf{0.23} &        \cellcolor{appearance}\textbf{0.06} & \cellcolor{appearance}\textbf{0.03} &     \cellcolor{appearance}\textbf{0.51} &       \cellcolor{appearance}\textbf{0.20} &      \cellcolor{appearance}\textbf{0.10} &       \cellcolor{appearance}\textbf{0.17} &    \cellcolor{appearance}0.84 \\ 
    \bottomrule
    \end{tabular}
    \,
    \begin{tabular}{l|ccc}
    \toprule
    {} & \multicolumn{3}{c}{Uniformity} \\
    \midrule
    {} &  Default &  \cellcolor{spatial}Spatial &  \cellcolor{appearance}Appearance \\
    \midrule
    Supervised &     3.93 &     3.87 &    3.84 \\
    Default    &     \textbf{3.84} &     3.83 &    3.67 \\
    \cellcolor{spatial_dark}Spatial    &     3.86 &     \cellcolor{spatial}\textbf{3.78} &    \underline{3.88} \\
    \cellcolor{appearance_dark}Appearance     &     \underline{3.98} &     \underline{3.98} &    \cellcolor{appearance}\textbf{3.61} \\
    \bottomrule
    \end{tabular}
    }
    \\
    \resizebox{1.0\linewidth}{!}{%
    \begin{tabular}{l|ccccccccccc}
    \toprule
    {} &  \cellcolor{spatial}Flickr1024 &  \cellcolor{spatial}COIL100 &  \cellcolor{spatial}ALOI &  \cellcolor{spatial}ALOT &  \cellcolor{grey}ALOI &  \cellcolor{grey}ALOT &  \cellcolor{appearance}ALOI &  \cellcolor{appearance}ALOT &  \cellcolor{appearance}ExposureErrors &  \cellcolor{appearance}RealBlur \\
    \midrule
    {} &  \cellcolor{spatial}Stereo &  \cellcolor{spatial}Pose/Scale &  \cellcolor{spatial}Viewp. &  \cellcolor{spatial}Viewp. &  \cellcolor{grey}Illumin. &  \cellcolor{grey}Illumin. &  \cellcolor{appearance}Temp. &  \cellcolor{appearance}Temp. &  \cellcolor{appearance}Exposure &  \cellcolor{appearance}Blur \\
    \midrule
    Random     &        0.75 &     1.17 &            1.04 &            1.51 &               0.85 &               1.13 &              0.20 &              0.55 &            1.19 &      0.17 \\
    Supervised &        0.20 &     0.60 &            0.72 &            0.66 &               0.47 &               0.51 &              0.06 &              0.21 &            0.33 &      0.22 \\
    Default    &        \textbf{0.12} &     0.49 &            0.46 &            0.60 &               0.33 &               0.48 &              0.04 &              0.07 &            0.20 &      0.20 \\
    \cellcolor{spatial_dark}Spatial    &        \cellcolor{spatial}\textbf{0.12} &     \cellcolor{spatial}\textbf{0.32} &            \cellcolor{spatial}\textbf{0.24} &            \cellcolor{spatial}\textbf{0.56} &               \textbf{0.20} &               \textbf{0.40} &              0.11 &              0.44 &            0.62 &      0.36 \\
    \cellcolor{appearance_dark}Appearance     &        0.52 &     0.92 &            0.73 &            1.37 &               0.41 &               0.78 &              \cellcolor{appearance}\textbf{0.01} &              \cellcolor{appearance}\textbf{0.02} &            \cellcolor{appearance}\textbf{0.09} &      \cellcolor{appearance}\textbf{0.04} \\
    \bottomrule
    \end{tabular}
    \,
    \begin{tabular}{l|ccc}
    \toprule
    {} & \multicolumn{3}{c}{ACR} \\
    \midrule
    {} &  Default & \cellcolor{spatial}Spatial & \cellcolor{appearance}Appearance \\
    \midrule
    Random     &     0.02 &     0.05 &    0.44 \\
    Supervised &     0.69 &     0.86 &    0.86 \\
    Default    &     \textbf{0.84} &     0.86 &    \textbf{0.90} \\
    \cellcolor{spatial_dark}Spatial    &     0.32 &     \cellcolor{spatial}\textbf{0.89} &    0.45 \\
    \cellcolor{appearance_dark}Appearance     &     0.35 &     0.35 &    \cellcolor{appearance}\textbf{0.90} \\
    \bottomrule
    \end{tabular}
    }
\end{table}

\subsection{Alternative metrics} \label{sec:alt_measure}
We note that there are alternative metrics for (in)variance such as alignment and uniformity \cite{Wang2020UnderstandingHypersphere} and ACR \cite{Wang2022ChaosLearning}. 
We provide results with these metrics here in Table \ref{tab:alignment}, and they lead to the same conclusions reported in the main paper.

\begin{figure*}
    \centering
    \includegraphics[width=1\linewidth]{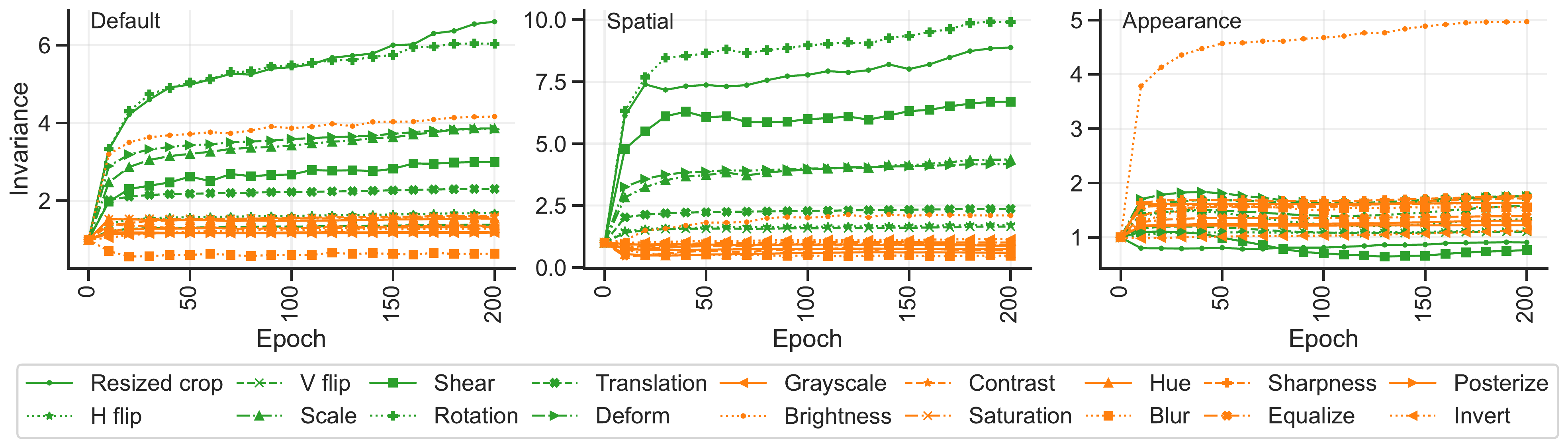}
    \caption{Invariances as measured by cosine similarity during pre-training. Different invariances are learned at different speeds. After 200 epochs many of the invariances are still steadily increasing, suggesting longer training helps achieve stronger invariances.}
    \label{fig:inv_pretraining}
\end{figure*}
\subsection{How do invariances change during pre-training?} 
It is clear from the results above that the use of augmentations leads to invariances to those transforms. But how do these invariances change as the model learns? Figure \ref{fig:inv_pretraining} shows how the invariances evolve during pre-training. The results echo those in Table~\ref{tab:synthetic_invariances}, showing the Appearance and Spatial models quickly specialise to greater corresponding invariances than the default model which has a moderate invariance to all transforms. In terms of the temporal dynamics, while some invariances stabilise quickly, other are continuing to increase at 200 epochs. This suggests that longer training may lead to further increases in invariance, and may explain why several state of the art learners achieve best performance with a very large number of iterations \cite{Caron2020UnsupervisedAssignments,Chen2020ARepresentations}.

\subsection{Do learned invariances hold for uncurated images?}

\begin{figure}[t]
    \centering
    \includegraphics[width=0.7\linewidth]{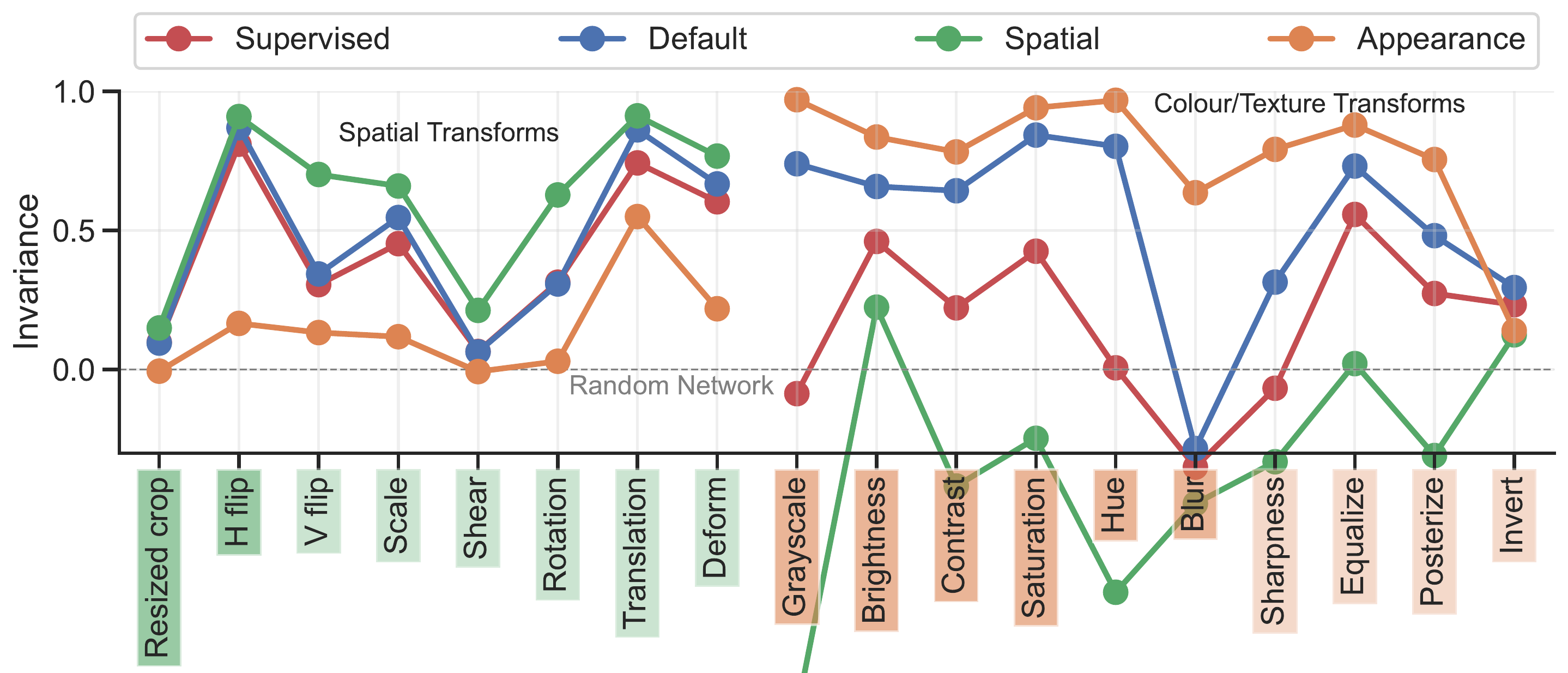}
    \includegraphics[width=0.7\linewidth]{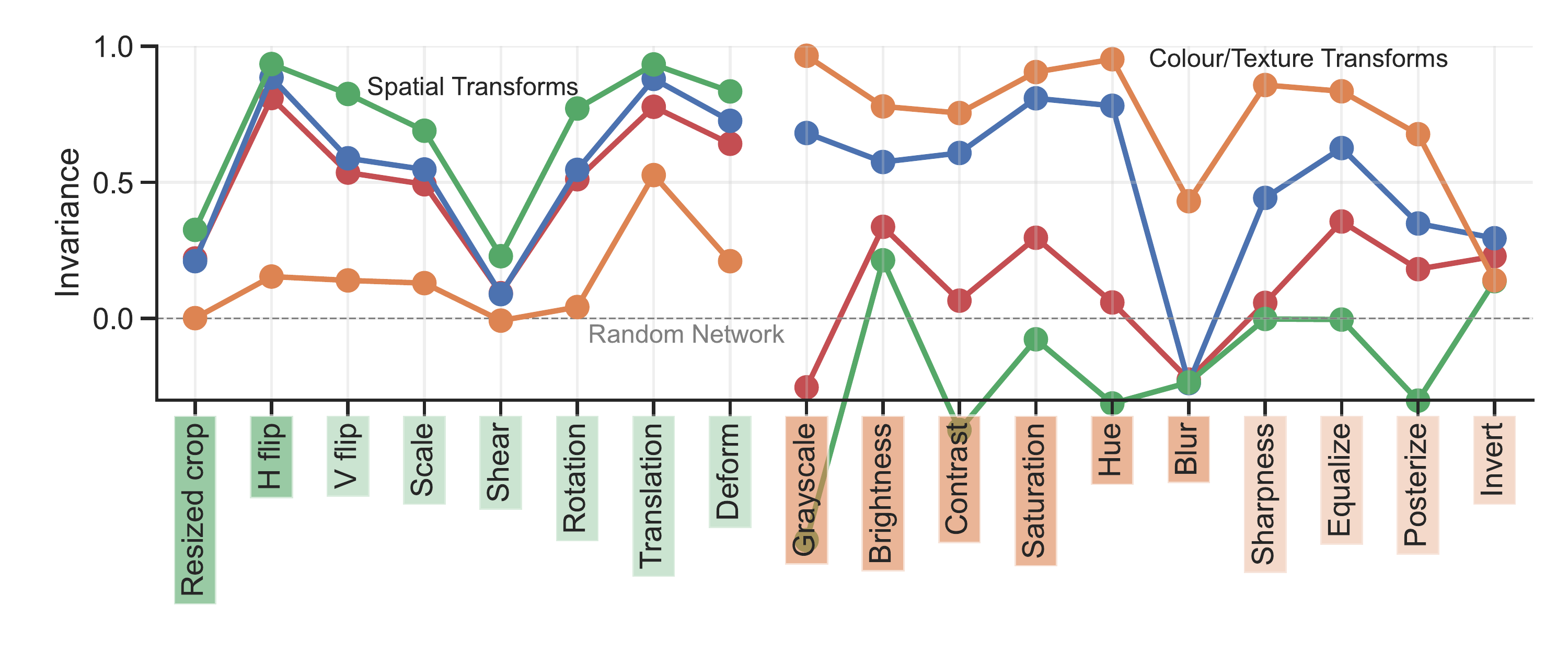}
    \caption{Invariances to synthetic transforms hold for cluttered images from COCO (top) and in-the-wild images from iNaturalist (bottom).}
    \label{fig:synthetic_invariances_coco_and_inat}
\end{figure}

We also evaluate the invariance to synthetic transforms on 100 images from the MS COCO val2017 set \cite{Lin2014MicrosoftContext} and iNaturalist 2021 validation set \cite{VanHorn2021BenchmarkingCollections}. As can be seen in Fig.~\ref{fig:synthetic_invariances_coco_and_inat}, the invariances of all models match those in Fig.~\ref{fig:teaser} (top) on ImageNet, showing that these learned invariances are not limited to highly curated object-centric images, but also to the cluttered images of COCO and the in-the-wild nature of iNaturalist.

\section{Datasets with real-world transforms} \label{sec:real_world_datasets}
Here we describe the datasets used in Section \ref{sub:intrinsic}:
\keypoint{Flickr1024} \cite{Scharstein2014High-ResolutionTruth} contains stereo image pairs. We use this to measure invariance to small lateral changes in viewpoint.
\keypoint{COIL-100} \cite{Nene1996ColumbiaCOIL100} is a dataset of 7,200 images of 100 objects photographed under varying pose conditions against a black background. Each object is imaged at every 5 degree angle of rotation around the vertical axis resulting in 72 images. This creates a suitable dataset for us to investigate the real-world viewpoint invariance of our models.
\keypoint{ALOI} \cite{Geusebroek2005TheImages} contains 1,000 objects captured under varying poses, camera angles and lighting conditions, allowing measurements of both viewpoint and illumination invariance. \keypoint{ALOT} \cite{Burghouts2009Material-specificFeatures} is a similar dataset in structure but contains photos of 250 textures instead of objects.
\keypoint{ExposureErrors} \cite{Afifi2021LearningCorrection} is composed of 24,000 images rendered at different exposures from raw RGB data.
\keypoint{RealBlur} \cite{Rim2020Real-WorldAlgorithms} has 4,500 geometrically aligned pairs of blurred and sharp images, thereby providing an ideal setup for measuring real-world blur invariance.

\section{Correlations}
\subsection{Between synthetic and real-world transforms}
\begin{figure}
    \centering
    \includegraphics[width=0.7\linewidth]{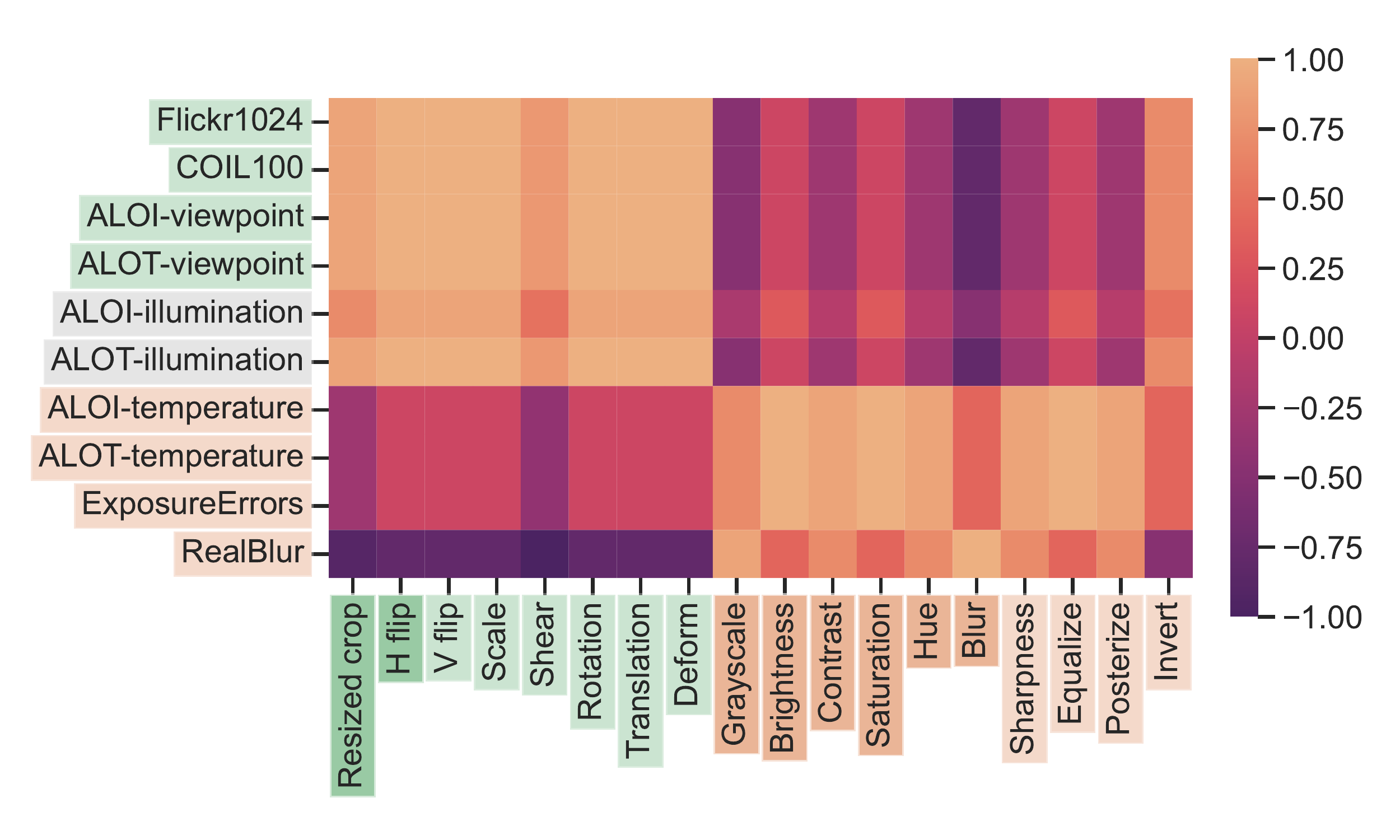}
    \caption{Rank correlation between invariances to synthetic and real-world transforms. Invariance to synthetic transforms is highly correlated with invariance to the corresponding family of real-world transforms.}
    \label{fig:rank_correlation}
\end{figure}

To establish the ability of synthetic transforms to capture the invariances found in the real world, we compute the correlations between our synthetic and real-world invariance measurements. In Figure \ref{fig:rank_correlation} we see that invariance to synthetic spatial transforms correlate very strongly with spatial real-world transforms. Furthermore, within the spatial family all transforms are correlated. These effects are similar for our appearance family. Interestingly, illumination changes on real-world data correlate strongly with spatial transforms like cropping and flipping. Intuitively, we would expect it to correlate stronger with synthetic changes in brightness. However, due to the very low illumination in the ALOI/ALOT data, only small regions of the objects are illuminated in each view, meaning that across views it can appear that the object is moving. An outlier in these results is the invert transform which correlates stronger with real-world spatial changes. These results suggest that many common synthetic transforms used in self-supervised pre-training lead to invariances that transfer to corresponding real-world transforms.

\subsection{Between synthetic and real-world transforms}
\begin{figure}[t]
    \centering
    \includegraphics[width=0.7\linewidth]{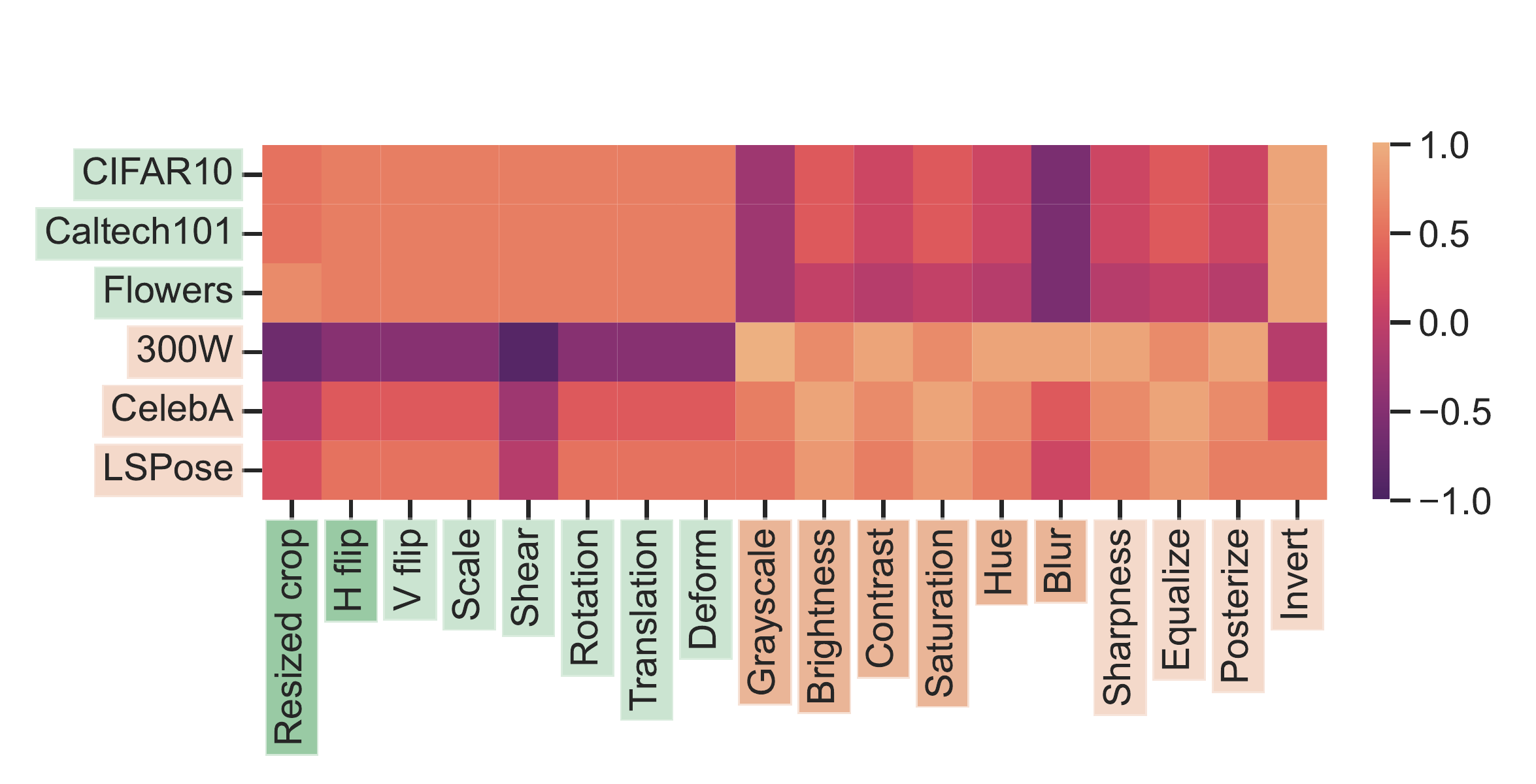}
    \caption{Rank correlation between synthetic invariances and downstream tasks of our set of models in Table~\ref{tab:downstream_tasks} (top). 
    A high value 
    means that learning a stronger invariance to the transform is highly correlated with getting better at the task.}
    \label{fig:corr_synthetic_invariance_downstream_task}
\end{figure}

Figure \ref{fig:corr_synthetic_invariance_downstream_task} shows the correlations between synthetic invariances and downstream model performances. We again see a block-like structure, where spatial invariances correlate with classification tasks and appearance invariances correlate with regression tasks. The task that stands out the most is 300W, where invariance to spatial transforms is very destructive for performance. The invert transform is also here an outlier as it correlates strongly with classification. 

\section{Downstream evaluation}

For all datasets, we use the full sets of images (combining train, val and test sets) for 5-fold cross-validation. For classification datasets we stratify the folds to ensure class balance. On Caltech101 and Flowers this means that we randomly select 30 and 20 images per class, respectively, to form the train set in the current fold and test on the rest. On 300W \cite{Sagonas2016300Challenge} we use images from both the indoor and outdoor sets. For CIFAR10 we report accuracy and for Caltech101 and Flowers, mean per-class accuracy. On 300W and CelebA we perform facial landmark regression and report the $R^2$ regression metric and for Leeds Sports Pose we perform pose estimation and report $R^2$.

We follow the evaluation of \cite{Chen2020ARepresentations}, but additionally perform 5-fold cross-validation. We extract features from a frozen backbone and, for classification datasets (CIFAR10, Caltech101, Flowers) fit a logistic regression classifier and for regression datasets (300W, CelebA and Leeds Sports Pose) we fit ridge regression. For both settings, the $\ell2$ hyperparameter search range is 45 logarithmically spaced values between $10^{-6}$ to $10^5$. For the larger and fused models we shift the range to $10^{-5}$ to $10^6$ for Spa+App and $10^{-4}$ to $10^7$ for Def+Spa+App, $3\times$Default and Default($\times 3$).

\begin{figure}[t]
    \centering
    \includegraphics[width=0.8\linewidth]{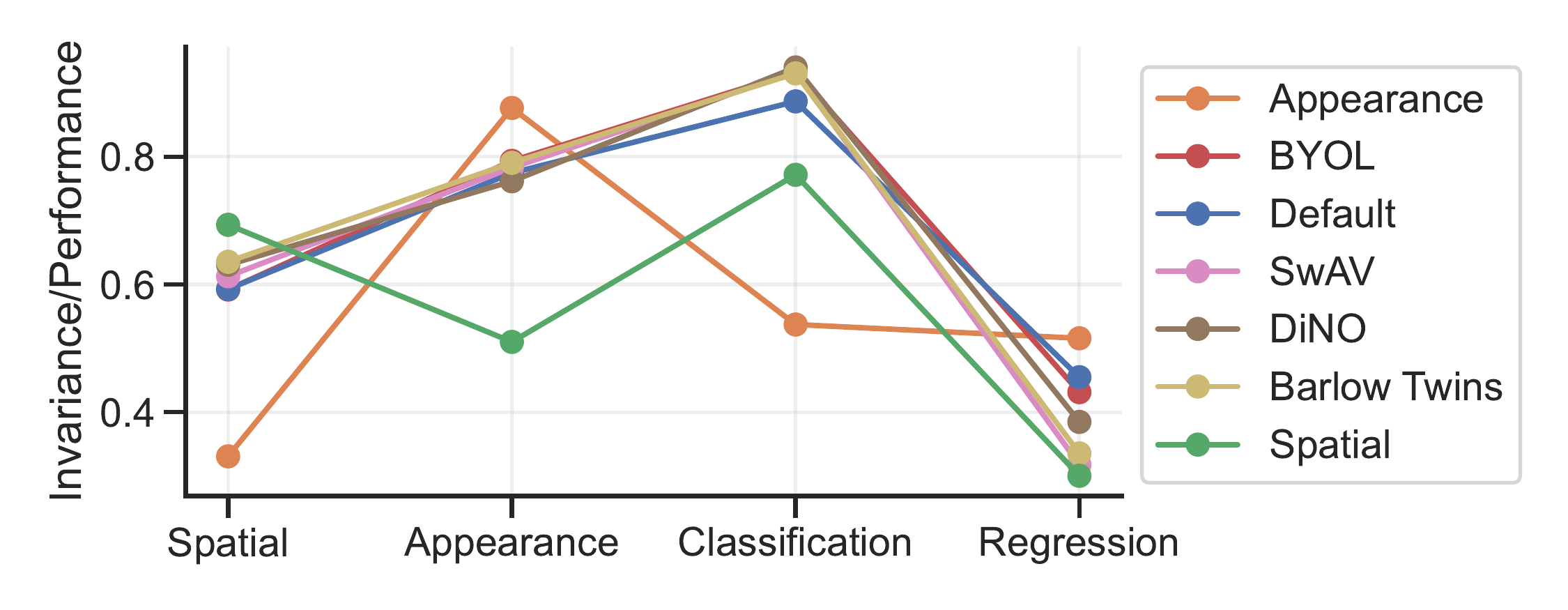}
    \caption{A parallel coordinates plot demonstrating the spatial/appearance invariance (measured in cosine similarity) and the performance on downstream tasks (measured in accuracy or $R^2$) for different existing state-of-the-art self-supervised models.}
    \label{fig:v_d_vs_downstream_existing_models}
\end{figure}

\subsection{Comparison of other State of the Art} 
Our experiment so far focused on pre-training a representative MoCo model with different augmentations. In the final experiment, we broaden our scope and evaluate a suite of existing pre-trained methods on our suite of tests for spatial and appearance invariances (Section~\ref{sec:synthetic}) and the downstream tasks studied in Section~\ref{sec:downstream_tasks}.  We evaluate BYOL \cite{Grill2020BootstrapLearning}, SwAV \cite{Caron2020UnsupervisedAssignments}, Barlow Twins \cite{Zbontar2021BarlowReduction} and DiNO \cite{caron2021emerging} along with our MoCo default, Spatial and Appearance models. All use a  ResNet50 backbone.

\keypoint{Results} From the plot in Figure~\ref{fig:v_d_vs_downstream_existing_models}, we can see that: (i) All the standard models fall between the performance of the Spatial and Appearance models in terms of spatial and appearance invariance (left two metrics). This shows that it is not possible to achieve high appearance and spatial invariance simultaneously. (ii) While the particular suite of default invariances has been well tuned for classification (with all existing methods substantively outperforming both our appearance and spatial models), it is poorly tuned for regression, where our Appearance model performs best. In fact, all other existing models shown here underperform MoCo-v2 on regression tasks, further highlighting the continuing overfitting of SOTA self-supervision to recognition tasks. It provides more evidence that the default augmentation suite has been overfitted to the most common benchmarks, and that more thought is necessary to design augmentation distributions suitable for more diverse downstream tasks.

\cut{
\section{Extra}
\begin{figure}
    \centering
    \includegraphics[width=\linewidth]{figures/rank_corr_downstream_all.pdf}
    \caption{Rank correlation between synthetic invariances and downstream tasks of models during pre-training. A high value 
    means that learning a stronger invariance to the transform is highly correlated with getting better at the task.}
    \label{fig:corr_synthetic_invariance_downstream_task_pretraining}
\end{figure}
}

\end{document}